\documentclass[3p,authoryear]{elsarticle}

\usepackage[utf8]{inputenc}
\usepackage{graphicx}%
\usepackage{multirow}%
\usepackage{amsmath,amssymb,amsfonts}%
\usepackage{amsthm}%
\usepackage{mathrsfs}%
\usepackage[title]{appendix}%
\usepackage{xcolor}%
\usepackage{textcomp}%
\usepackage{manyfoot}%
\usepackage{booktabs}%
\usepackage{algorithm}%
\usepackage{algorithmicx}%
\usepackage{algpseudocode}%
\usepackage[normalem]{ulem}
\usepackage{subcaption}
\usepackage{enumitem}
\usepackage{flafter}
\usepackage[hidelinks,colorlinks=true]{hyperref}
\usepackage{underscore}% For underscore in file and attribute names of GTFS
\usepackage{orcidlink}%
\usepackage{listings, lstautogobble}
\usepackage[most]{tcolorbox}
\usepackage{booktabs}
\usepackage[T1]{fontenc}
\usepackage{colortbl}
\usepackage{tikz}
\usepackage{array}
\usetikzlibrary{positioning, shapes.misc, calc, arrows.meta, shapes.geometric, arrows, fit}
\usepackage{makecell}
\setcellgapes{4pt}%parameter for the spacing

\definecolor{dkgreen}{cmyk}{0.65, 0.3, 0.8, 0.2}
\definecolor{gray}{rgb}{0.5,0.5,0.5}
\definecolor{mauve}{rgb}{0.58,0,0.82}
\definecolor{lavender}{cmyk}{0.4,0.6,0,0}
\definecolor{dkblue}{cmyk}{1, 0.5, 0, 0.5}
\definecolor{darkred}{cmyk}{0, 1, 1, 0.2}
\definecolor{darkgreen}{cmyk}{1, 0, 1, 0.5}
\definecolor{darkblue}{cmyk}{1, 0.5, 0, 0.5}
\definecolor{darkpurple}{cmyk}{0.5, 1, 0, 0.5}
\definecolor{darkbrown}{cmyk}{0.5, 1, 1, 0.5}
\definecolor{stringColor}{rgb}{0.5, 0.6, 0.2}
\definecolor{darkorange}{RGB}{205, 102, 0}
\definecolor{delim}{RGB}{20,105,176}
\definecolor{numb}{RGB}{106, 109, 32}
\definecolor{string}{rgb}{0.64,0.08,0.08}

\lstdefinelanguage{PythonCustom}{
  language=Python,
  sensitive=true,
  morestring=[b]",
  morestring=[b]',
  stringstyle=\color{dkgreen},
}

\lstdefinelanguage{json}{
    showspaces=false,
    showtabs=false,
    breaklines=true,
    postbreak=\raisebox{0ex}[0ex][0ex]{\ensuremath{\color{gray}\hookrightarrow\space}},
    breakatwhitespace=true,
    basicstyle=\ttfamily\small,
    upquote=true,
    tabsize=2,
    morestring=[b]",
    stringstyle=\color{string},
    literate=
     *{0}{{{\color{numb}0}}}{1}
      {1}{{{\color{numb}1}}}{1}
      {2}{{{\color{numb}2}}}{1}
      {3}{{{\color{numb}3}}}{1}
      {4}{{{\color{numb}4}}}{1}
      {5}{{{\color{numb}5}}}{1}
      {6}{{{\color{numb}6}}}{1}
      {7}{{{\color{numb}7}}}{1}
      {8}{{{\color{numb}8}}}{1}
      {9}{{{\color{numb}9}}}{1}
      {\{}{{{\color{delim}{\{}}}}{1}
      {\}}{{{\color{delim}{\}}}}}{1}
      {[}{{{\color{delim}{[}}}}{1}
      {]}{{{\color{delim}{]}}}}{1},
}

\newcommand{\backtick}[1]{%
  \textcolor{dkblue}{%
    \colorbox{gray!10}{%
      \ttfamily\frenchspacing#1%
    }%
  }%
}

\newcommand{\backtickRed}[1]{%
  \textcolor{darkred}{%
    \colorbox{gray!10}{%
      \ttfamily\frenchspacing#1%
    }%
  }%
}

\definecolor{humanTitleColor}{RGB}{230,230,230}  % Light gray for human title
\definecolor{messageColor}{RGB}{255,255,255}     % White for message background
\tcbset{
human/.style={
enhanced,
arc=1mm,
fuzzy shadow={0mm}{-0.5mm}{0mm}{0.1mm}{black!20},
title={Human:},
fonttitle=\small\bfseries,
coltitle=gray!30!black,
attach boxed title to top left={yshift=-1.5mm, xshift=3mm},
boxed title style={size=small, colback=white},
before skip=1.5pt,
after skip=3pt,
boxsep=1.5pt,
fontupper=\footnotesize,
fontlower=\footnotesize,
left=1.5pt,
right=1.5pt,
bottom=1.5pt,
boxrule=1pt,
},
llm/.style={
enhanced,
colback=messageColor,
arc=1mm,
fuzzy shadow={0mm}{-0.5mm}{0mm}{0.1mm}{black!20},
title={TransitGPT:},
fonttitle=\small\bfseries,
coltitle=black!80,
colframe = black!80,
attach boxed title to top left={yshift=-1.5mm, xshift=3mm},
boxed title style={size=small, colback=white},
before skip=1.5pt,
after skip=1.5pt,
boxsep=1.5pt,
fontupper=\footnotesize,
fontlower=\footnotesize,
left=3pt,
right=3pt,
bottom=1.5pt,
boxrule=1pt,
},
prompt/.style={
enhanced,
arc=2mm,
fonttitle=\small,
fuzzy shadow={0mm}{-0.5mm}{0mm}{0.2mm}{black!30},
attach boxed title to top left={yshift=-2mm, xshift=2mm},
boxed title style={size=small, colback=black!80},
before skip=4pt,
after skip=4pt,
boxsep=4pt,
fontupper=\small,
fontlower=\small,
bottom=2pt,
}
}

\journal{Public Transport}

\newcounter{stepcounter}
\newcommand{\step}[1]{%
    \vspace{1em} % Add some space before the step
    \refstepcounter{stepcounter} % Increment the step counter
    \noindent\textbf{Step \thestepcounter: #1} % Format the step title
    \vspace{.5em}
    \par % Start a new paragraph
}

\begin{document}
\begin{frontmatter}

    
    %% use the read command for the email address,
    
    \title{TransitGPT: A Generative AI-based framework for interacting with GTFS data using Large Language Models} %% Article title

    
    \author[label1]{Saipraneeth Devunuri\corref{cor1} \orcidlink{0000-0002-5911-4681}} %% Author name
    \author[label1]{Lewis Lehe \orcidlink{0000-0001-8029-1706}} %% Author name
    \affiliation[label1]{organization={Department of Civil Engineering, University of Illinois at Urbana Champaign},%Department and Organization
        city={Urbana},
        state={IL},
        country={USA}}

    \cortext[cor1]{Corresponding author: Saipraneeth Devunuri, \url{sd37@illinois.edu}}
    
    % \affiliation{organization={},%Department and Organization
    
    %% Abstract
    \begin{abstract}
        This paper introduces a framework that leverages Large Language Models (LLMs) to answer natural language queries about General Transit Feed Specification (GTFS) data. The framework is implemented in a chatbot called TransitGPT with open-source code. TransitGPT works by guiding LLMs to generate Python code that extracts and manipulates GTFS data relevant to a query, which is then executed on a server where the GTFS feed is stored. It can accomplish a wide range of tasks, including data retrieval, calculations and interactive visualizations, without requiring users to have extensive knowledge of GTFS or programming. The LLMs that produce the code are guided entirely by prompts, without fine-tuning or access to the actual GTFS feeds. We evaluate TransitGPT using GPT-4o and Claude-3.5-Sonnet LLMs on a benchmark dataset of 100 tasks, to demonstrate its effectiveness and versatility. The results show that TransitGPT can significantly enhance the accessibility and usability of transit data. 
    \end{abstract}
    
    
    % \begin{highlights}
    % \item This work presents a novel LLM architecture that integrates with existing python libraries and can handle a wide variety of tasks related to GTFS data retrieval and analysis.
    % \item We utilize several prompt engineering techniques that LLMs to understand transit data better thereby enhancing the performance of LLMs in transit domain.
    % \item We built TransitGPT, an open-source chatbot which provides a user interface to retrieve transit information through natural language instructions. The interface also provides a well-documented code that helps with code assistance and further analysis.
    % \item A benchmark dataset consisting of 100 tasks related to GTFS data retrieval tasks is introduced, covering 8 different task categories and spanning various complexity levels. 
    % \end{highlights}
    
    %% Keywords
    
\end{frontmatter}



\section{Introduction}\label{sec1}
%% Labels are used to cross-reference an item using \ref command.

\subsection{Background}

The General Transit Feed Specification (GTFS) is an open data standard for transit data. Started in 2005 as a collaboration between Google and TriMet (Portland, OR's transit agency) \citep{Roth2010, McHugh2013}, today over 10,000 agencies from over 100 countries use GTFS to share and store their transit feeds publicly \citep{MobilityData2024}. The scope of GTFS has expanded beyond schedules to include information about real-time updates, flexible services, and fares. Its widespread adoption has led to the development of software, programming libraries, and plugins for popular GIS software that help agencies create or visualize GTFS feeds.

The principal use of GTFS is for navigation applications such as Google Maps, Apple Maps, Transit, etc. But in addition to helping people \emph{use} transit, GTFS data is increasingly used to \emph{analyze}, \emph{measure} and \emph{understand} transit systems. A spatio-temporal decomposition of GTFS has been utilized to construct O-D travel time matrices, generate isochrones, and estimate travel time uncertainty \citep{Pereira2021, Liu2024}. This decomposition has been used to conduct transit accessibility analysis by examining socioeconomic and spatial-temporal inequalities \citep{Fayyaz2017, Prajapati2020, Yan2022, Pereira2023}. \citet{Devunuri2024b,Devunuri2024d} decomposes GTFS feeds into segments for all US and Canadian transit agencies to measure statistics about bus stop spacings. \citet{Fortin2016} modeled the transit timetables as graphs and developed graph-oriented indicators, including connectivity between stops and the proportion of active stop pairs over time. GTFS has also been used to create informative visualizations. \citet{Kunama2017} developed `GTFS-Viz', a tool to animate the movement of buses with time and visualize the number of buses in use. \citet{Para2024} created a user-friendly interface to visualize speeds, service rates, and headways. In conjunction with GTFS-Realtime, \citet{Park2020} examined delay and its propagation in the context of transit system performance.

One challenge faced in analyzing GTFS data is that the specification itself is complex. A feed may include over 30 .txt files containing over 200 fields connected in complex ways. Some fields are required, some are optional, and others are \emph{conditionally required or forbidden} (i.e., only included or excluded if another optional field is present). Feeds also include a variety of qualitatively different types of data---including times, stop coordinates, route shapes, identifier strings, integer counts, fare rules, fare amounts, and color hexadecimal. Thus, to analyze a system, the user must appreciate the relationships among these fields. \citet{Devunuri2024f} shows that two popular Large Language Models (LLMs), GPT-3.5-Turbo and GPT-4, can ameliorate this challenge somewhat by answering questions about the GTFS specification with reasonable accuracy. A second challenge is that to obtain specific information about an agency (or agencies) from its GTFS feed, an analyst (even one who understands the GTFS specification well) must also figure out how to read and manipulate the data with a spreadsheet application or some computer code in order to obtain the answers sought. To this end, \citet{Devunuri2024f} also briefly explores using LLMs to write code that carries out simple retrieval tasks on GTFS data---an application of what is known as \emph{program synthesis} \citep{Haluptzok2023}.

\subsection{Plan of paper}

\begin{figure}[!ht]
    \centering
    \begin{subfigure}[b]{0.45\textwidth}
        \centering
        \begin{tcolorbox}[human]
            What is the shortest route in the feed?
        \end{tcolorbox}
        \begin{tcolorbox}[llm]
            The shortest route in the feed is \backtick{RUBY_EVENING} with a length of \backtick{1.6km}. Here are some additional details:
            \begin{itemize}
                \begin{footnotesize}
                    \item Route ID: \backtick{RUBY_EVENING}
                    \item Route Short Name: \backtick{110-RUBY_EVENING}
                    \item Route Type: \backtick{3}
                    % \item Route Long Name: \backtick{Ruby Evening}
                    % \item Length in Meters: \backtick{1654.46 meters}
                \end{footnotesize}
            \end{itemize}
        \end{tcolorbox}
        \caption{Text Information about the shortest route [CUMTD]}
    \end{subfigure}
    \hfill
    \begin{subfigure}[b]{0.52\textwidth}
        \centering
        \begin{tcolorbox}[human]
            Which services and their respective routes are operational today
        \end{tcolorbox}
        \begin{tcolorbox}[llm]
            \resizebox{\textwidth}{!}{%
                \begin{tabular}{|c|c|}
                    \hline
                    \textbf{service\_id} & \textbf{route\_id}                                          \\
                    \hline
                    B2\_MF               & \backtick{BLUE}, \backtick{GOLD}                            \\
                    B1\_MF               & \backtick{GREEN}, \backtick{GREEN_EXPRESS}, \backtick{BLUE} \\
                    9A\_SHOW\_HSEO       & \backtick{GREEN_EXPRESS}                                    \\
                    \ldots               & \ldots                                                      \\
                    
                    \hline
                \end{tabular}
            }
        \end{tcolorbox}
        \caption{DataFrame consisting service information [CUMTD]}
    \end{subfigure}
    
    \hfill
    \begin{subfigure}[b]{0.4\textwidth}
        \centering
        \begin{tcolorbox}[human]
            Show me a map of stops on Market St
        \end{tcolorbox}
        \begin{tcolorbox}[llm]
            Sure, here is a map of all stops on Market St
            
            \includegraphics[width=\textwidth]{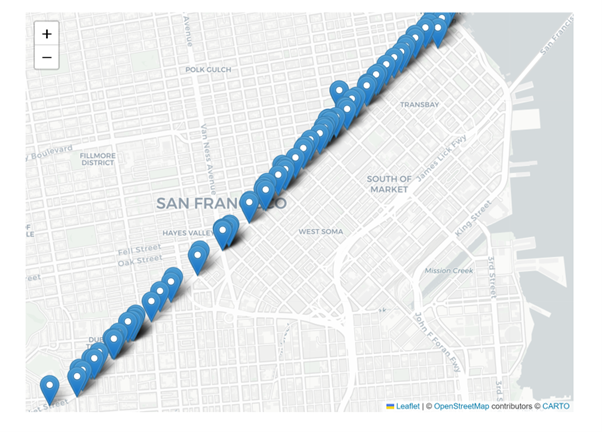}
        \end{tcolorbox}
        \caption{Map of stops on Market St [SFMTA]}
    \end{subfigure}
    \hfill
    \begin{subfigure}[b]{0.52\textwidth}
        \centering
        \begin{tcolorbox}[human]
            Plot histogram of segment speeds along with kernel density for the route \backtick{MALCOLM X/MAPLE}
        \end{tcolorbox}
        \begin{tcolorbox}[llm]
            \includegraphics[width=\textwidth]{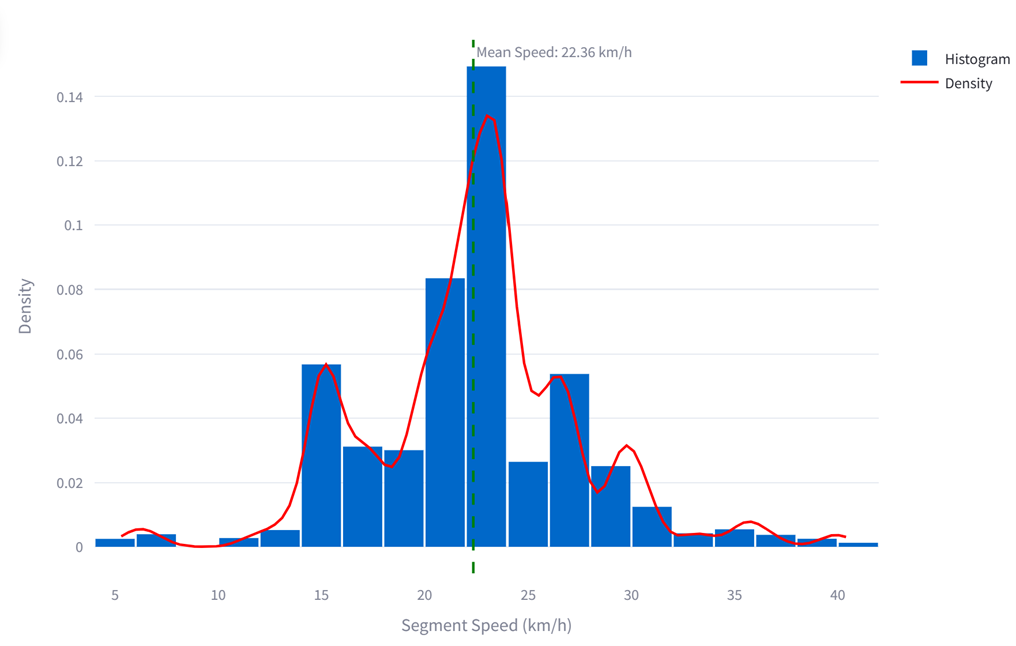}
        \end{tcolorbox}
        \caption{Histogram of segment speeds for a route [DART]}
    \end{subfigure}
    \caption{Demonstrations of  `TransitGPT' in generating responses for GTFS data retrieval tasks. Sample visualizations generated using TransitGPT are available in \ref{sec:sampleVisualizations}.}
    \label{fig:TransitGPT}
\end{figure}

This paper introduces an architecture that uses LLMs to answer questions about transit systems from GTFS data. This architecture is implemented in a chatbot interface called \textbf{TransitGPT}. The link to the chatbot as well as the code behind the TransitGPT is available on our GitHub repository located at \url{https://github.com/UTEL-UIUC/TransitGPT}. The reader is encouraged to experiment with TransitGPT by selecting one of the ten agencies we have prepared and asking questions about it. Figure \ref{fig:TransitGPT} shows some examples of questions and answers generated by TransitGPT. TransitGPT can answer questions that a rider might have, such as: ``When does the last Orange bus arrive at University and Victor on a Tuesday?'' It can also answer system-level questions such as: ``Which routes have headway shorter than 15 minutes on the weekend?'' 

A goal of this paper is to explain the architecture behind TransitGPT and how it was built. In a nutshell, what we do is (i) ask the LLM to write Python code that can answer a question given that the GTFS feed is available; then (ii) run that code on a server where the feed is stored. While this two-step combination conveys the gist, TransitGPT involves many embellishments which are explained more thoroughly below. In particular, we have created a large \emph{system prompt}: a text file containing instructions that guide the LLM as it writes code to answer the questions. This system prompt can be passed to any public-facing LLM, without `pre-training' or `fine-tuning'. This emphasis on prompt engineering (rather than fine-tuning) aligns with the concept of ``Prompting Is All You Need'' \citep{Feng2024}, which avoids the need for extensive training data and compute resources. To evaluate different prompt designs and to compare the performance of LLMs, we employ a 100-question benchmark. 

At the outset, it is important to distinguish our approach from another one which is perhaps more intuitive: passing an entire GTFS feed to an LLM and asking questions about the data. One reason we do not take this direction is that most GTFS feeds are too large to pass to the popular, public-facing LLMs currently due to context length limitations \citep{Hsieh2024}. Ostensibly, one could use Retrieval Augmented Generation (RAG) \citep{Lewis2021, Gao2024} to retrieve the feed in chunks, but an obstacle to this approach is the inter-related nature of GTFS files (i.e., there is no single file which contains all information about, say, a route). In either case, a more fundamental problem is that many queries require not only looking up data but also manipulating data, doing calculations with data, and connecting different pieces of data according to the complex rules of GTFS. While LLMs excel at tasks such as translation, summarization, and classification, they often struggle with tasks that need precise numerical calculations or data manipulation \citep{Ahn2024}. For instance, it has been documented that LLMs make mistakes with simple tasks such as counting the number of r's in the word `strawberry' or deciding whether `9.11' or `9.7' is a larger number. By contrast, it is straightforward for an LLM to write reliable code that solves those problems.

The paper is organized as follows: 
\begin{itemize}[itemsep=2pt]
    \item Section \ref{relatedWork} presents relevant literature on LLM applications in code generation, transportation, and transit.
    \item Section \ref{sec:architecture} describes the approach and architecture of TransitGPT.
    \item Section \ref{sec:benchmark} evaluates the performance of TransitGPT on a benchmark dataset of 100 tasks which have been classified into \emph{eight} categories. We also compare the performance of TransitGPT with two state-of-the-art (SOTA) LLMs: GPT-4o \citep{OpenAI2024} from OpenAI and Claude-3.5-Sonnet \citep{Anthropic2024} from Anthropic.
    \item Section \ref{sec:conclusion} concludes with a discussion on the capabilities, limitations, and future directions of this work.
\end{itemize}

\subsection{Scope and contribution}

\begin{figure}[ht]
    \centering
    \includegraphics[width=.65\textwidth]{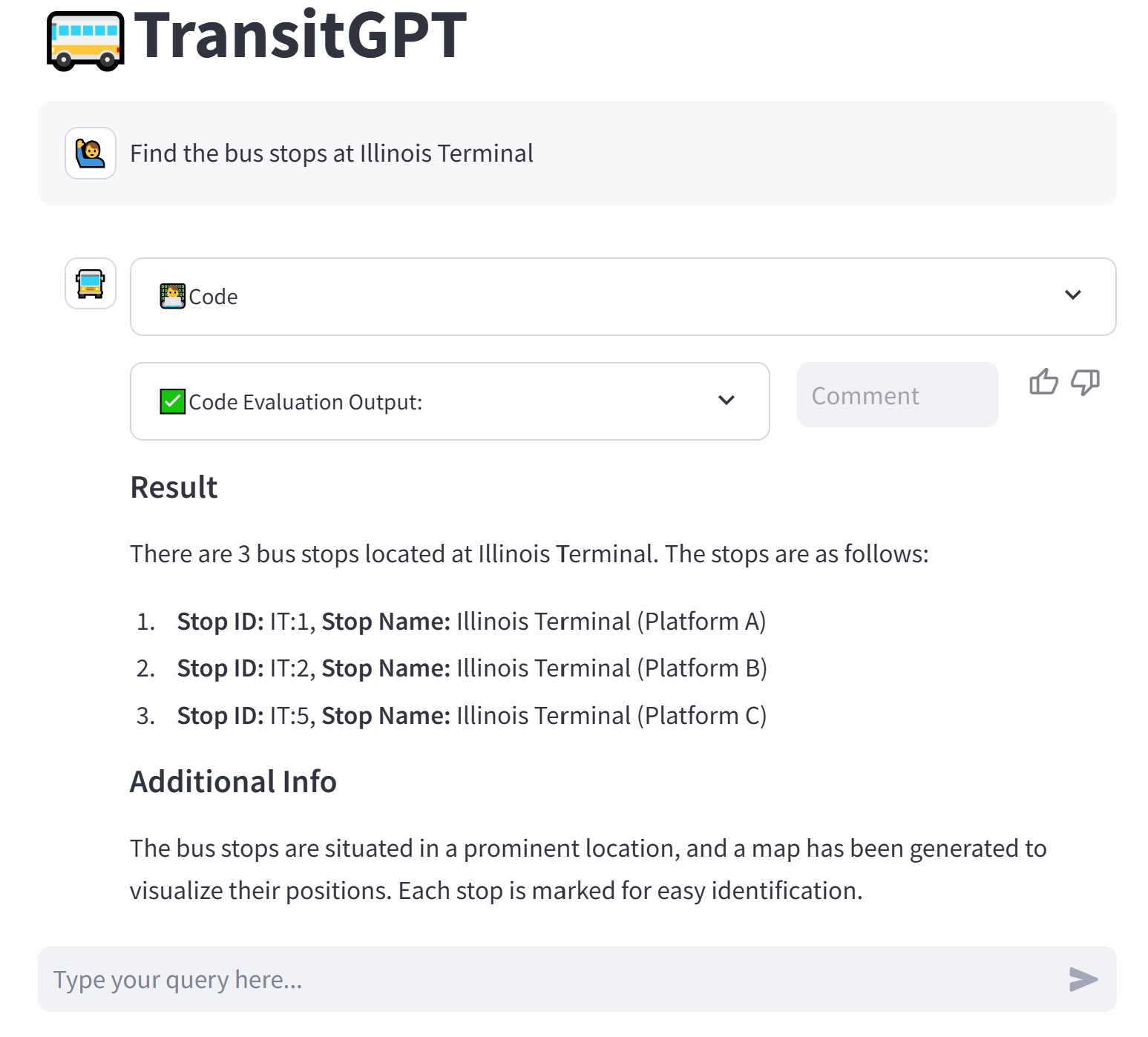}
    \caption{TransitGPT Interface}\label{fig:interface}
\end{figure}

Note the following limitations on the scope of the project: 
\begin{enumerate}[label=(\roman*)]
    \item The scope is limited to the \emph{GTFS Static} version of GTFS which conveys information that is planned in advance (e.g., stop locations, routes, schedules, etc.). TransitGPT does not draw information from GTFS Realtime feeds (about delays, vehicle locations, etc.).
    \item TransitGPT cannot generally answer questions that go beyond the capabilities of GTFS Static. For example, GTFS does not say how many seats are on each bus, so TransitGPT cannot answer questions about that. Geocoding is an exception; TransitGPT can use the Google Maps Geocoding API and the geocoding library Nominatim to answer queries such as ``How many stops are within 30 meters of Fisherman's Wharf?''
    \item Errors or limitations in a GTFS feed (which is prepared by a transit agency) will be reflected in the answers generated by TransitGPT. \citet{Devunuri2024c} surveys various errors (defined as deviations from GTFS standards) in GTFS Static feeds across the US and Canada and finds they are relatively common---although few of them are substantial. In addition to errors, many GTFS features are optional, so TransitGPT cannot answer questions that rely on an optional file (such as \texttt{fare_rules.txt} and \texttt{transfers.txt}) or features (such as the \emph{wheelchair_accessible} or  \emph{bikes_allowed} fields in \texttt{trips.txt}) that an agency has declined to include.
    \item TransitGPT is not intended for routing questions: e.g., ``How do I get from Chicago Union Station to the Harold Washington Library?'' This is simply not a space that TransitGPT meant to ``compete'' in. Mobile apps, such as Transit, are already optimized for routing: they utilize real-time data and forecasting which we do not access, and they have interfaces that show users when and where to walk (and wait/transfer) as well as choices among different options. TransitGPT's interface, shown in Figure \ref{fig:interface}, is designed for answering questions about systems. 
\end{enumerate}

The paper's contributions are intended to go beyond the current version of the TransitGPT chatbot itself. Narrowly, the 100-question benchmark is suitable for evaluating other approaches to the problem of making transit knowledge more accessible, and readers may think of other embellishments beyond those we describe that would make a transit chatbot more useful, accurate, or efficient. To this end, the code is open source. More broadly, it is hoped that the architecture and the prompt engineering behind it can inspire other applications of LLMs in the field of transportation. Transportation today is a domain that enjoys an increasing buffet of standardized data sources. As an example, the GTFS concept has inspired a General Bikeshare Feed Specification (GBFS)  which is an open data standard designed not only for bike-share but also other shared mobility services including scooters, mopeds, and cars \citep{MobilityData2024a}. The architecture of TransitGPT could be altered to answer queries about GBFS data.

\section{Related Work}\label{relatedWork}

\subsection{LLMs for Code Generation}
The ability of Large Language Models (LLMs) to generalize and learn from context (in-context learning) has been effectively utilized for coding tasks, including code completion\footnote{Code completion refers to the automatic completion of code segments, which aids in writing code, fixing errors, and providing in-line suggestions.} and code generation. While these tasks are complementary, this paper solely focuses on code generation. In the domain of data-related applications, Text2SQL \citep{Khatry2023,Zhang2024} has garnered significant attention for its ability to perform data extraction and analysis. Text2SQL allows users to query relational databases using natural language, which is then translated into SQL queries and executed. However, SQL lacks integration with existing libraries and does not offer visualization capabilities. Recently research has shown that Text2SQL performs poorly on tasks that require semantic or complex reasoning \citep{Biswal2024}.

Researchers have explored alternative approaches of using programming languages such as Python \citep{Haluptzok2023, Liu2023c} or Java \citep{Feng2024} and utilizing libraries written on top of them. \citet{Zan2022} demonstrated LLM's ability to generate code using popular data processing libraries numpy and pandas (using NumpyEval and PandasEval benchmarks) which are present in the corpus of training data of most LLMs. Furthermore, \citet{Patel2024} demonstrated that LLMs can learn and utilize libraries not present in their pre-training data through descriptions and raw code examples. This capability significantly expands the potential applications of LLMs in specialized domains such as transit, allowing the use of domain-specific libraries and tools. The ability to use libraries has also allowed LLMs to tackle much more complex tasks in related disciplines such as Data Science \citep{Zhang2024e} and Machine Learning \citep{Zhang2024b}.

\subsection{LLMs for Transportation}

The application of Large Language Models (LLMs) has expanded across various domains, including transportation. In the field of traffic safety, \citet{Zheng2023} investigated the potential of LLMs for automating accident reports and enhancing object detection through augmented traffic data. Similarly, \citet{Mumtarin2023} conducted a comparative analysis of LLMs in extracting and analyzing crash narratives, further demonstrating their utility in safety-related tasks. LLMs have also shown promise in autonomous vehicle technology. Research by \citet{Zhang2024a}, \citet{Cui2023} and \citet{Fu2023} explored the use of LLMs to analyze, interpret, and reason about various driving scenarios, potentially improving the decision-making capabilities of self-driving systems. In traffic management, LLMs have been applied to several critical tasks. \citet{Zhang2024d}, \citet{Movahedi2024} and \citet{Da2023} investigated the use of LLMs for signal control optimization, while \citet{Zhang2024f} explored their application in demand forecasting. These studies highlight the versatility of LLMs in addressing complex transportation challenges, from infrastructure management to predictive analytics.

\subsection{LLMs for Transit}
Applications of LLMs to public transit have been more limited but have grown in the past year. \citet{Devunuri2024f} developed benchmarks to evaluate LLMs' capabilities (zero-shot and few-shot) and limitations in understanding GTFS semantics and retrieving information from GTFS data. The `GTFS Semantics' benchmark tests LLMs on various GTFS elements, including terminology, data structures, attribute and category mapping, and common reasoning tasks. The `GTFS Retrieval' benchmark assesses LLMs' performance on tasks ranging from simple lookups to complex operations involving sorting, grouping, joining, and filtering information. In related work, \citet{Wang2024} employed LLMs for transit information extraction and sentiment analysis from social media platforms. They also demonstrated that Retrieval Augmented Generation (RAG) improved the performance of LLMs on GTFS Semantics and Retrieval benchmarks. \citet{Syed2024} developed TransportBench, a comprehensive benchmark comprising questions from transportation planning, economics, networks, geometric design, and transit systems. \citet{Oliveira2024} developed a methodology to investigate and evaluate bus stop infrastructure and its surroundings using street-view images and automated image descriptions from LLMs.

\section{Control Flow}\label{sec:architecture}

This section describes how TransitGPT works as a series of steps that control the flow of information. Every step takes place on a remote server hosted by \emph{Streamlit}, which is a company that provides a platform for building and sharing AI-centered web apps. The steps are executed by Python scripts on the Streamlit server. The control flow is described in broad strokes rather than in exact detail, but the reader may visit the documentation and code for TransitGPT to see exactly how things are carried out.

\begin{figure}[ht]
    \centering
    \includegraphics[width=\textwidth]{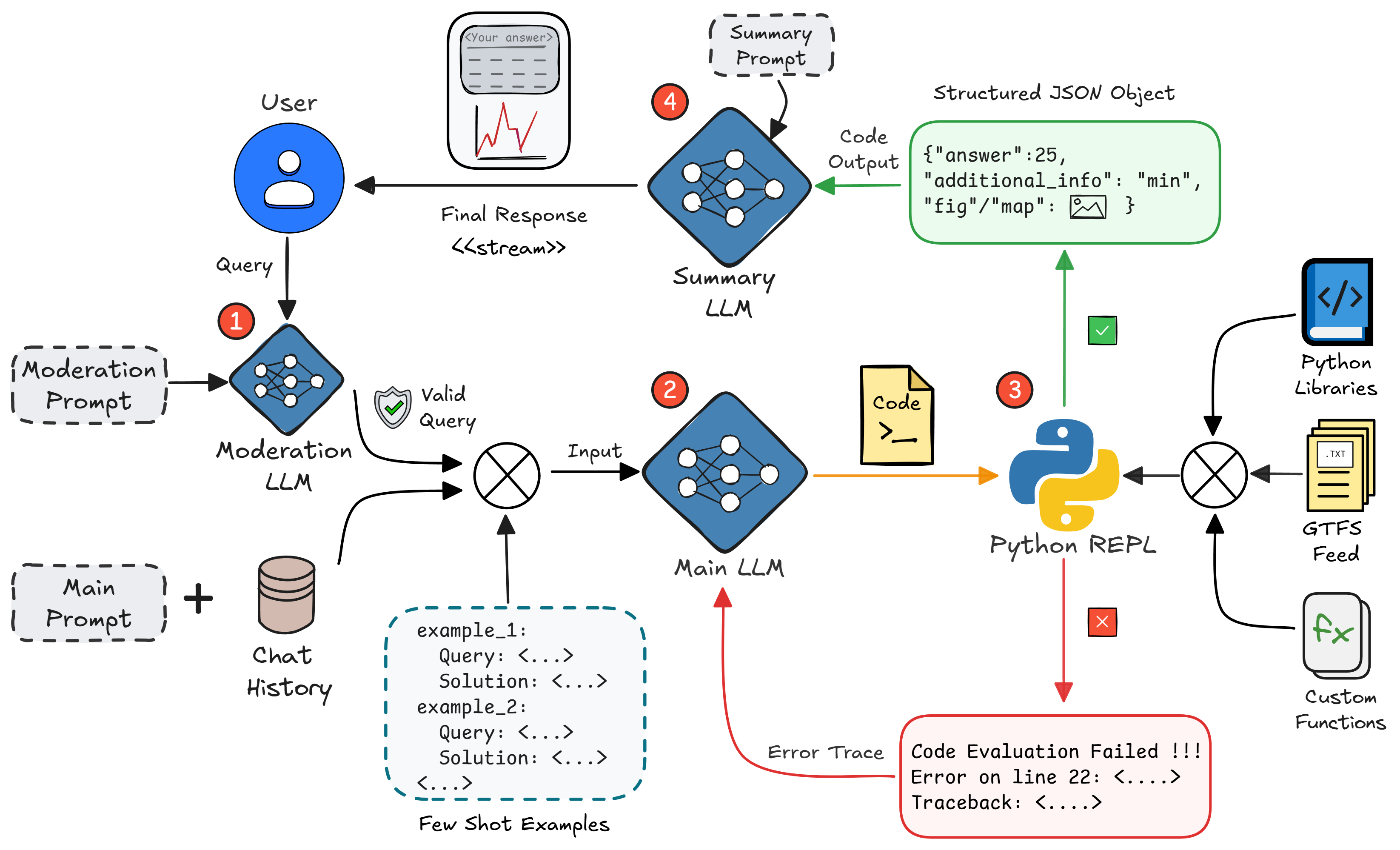}
    \caption{Extended TransitGPT Architecture}\label{fig:architectureExtended}
\end{figure}
\step{Pre-Query}\label{step:preQuery}

Before typing in a query (i.e., a question or command), the user selects a GTFS feed (e.g., SFMTA) and an LLM (e.g., GPT-4o) from among our ten agencies and four LLMs. 

For each GTFS feed, there is already a ``pickled'' (a serialized\footnote{Note it would be possible to load the raw GTFS feed into the sandbox, but the feeds are pre-prepared to save time when the user chooses an agency to investigate.} object saved to disk) \texttt{Feed} object which sits on the remote Streamlit server. A \texttt{Feed} is a Python object whose attributes correspond to the .txt files in a GTFS feed. For example, if \texttt{feed} is a \texttt{Feed} instance, then  \texttt{feed.routes} has data from the file \texttt{routes.txt}. Each .txt file's data is stored as a pandas DataFrame whose columns are the fields in the .txt file. Hence, \texttt{feed.routes.route\_id} yields a list of the strings given in the \texttt{route\_id} column of \texttt{routes.txt}. The Feed objects were created using the Python library \backtick{gtfs\_kit}. \ref{sec:preProcessing} details how the data from the actual GTFS feeds are pre-processed.

A Python program we call the ``sandbox'' runs constantly on the Streamlit server, waiting for pieces of code to execute. When the user selects a GTFS feed, the sandbox will import the corresponding (pickled) Feed object from disk to create a variable named \texttt{feed}. The sandbox also has imported eight  Python libraries\footnote{The libraries are \backtick{gtfs\_kit}, \backtick{pandas}, \backtick{numpy}, \backtick{geopandas}, \backtick{geopy}, \backtick{poltly.express}, \backtick{thefuzz}, and \backtick{folium}.} which have been downloaded onto the Streamlit server. 

\step{Moderation}\label{step:moderation}

Next, the user types a question or command into the TransitGPT interface: e.g., ``Show all the stops on Market St.'' We immediately send this query along with a document called the Moderation Prompt (Figure \ref{fig:moderationPrompt}) to GPT-4o-mini (a faster and cheaper version of GPT-4o). The Moderation Prompt instructs GPT-4o-mini to judge whether the user's query is relevant to transit. For example, the user cannot ask ``How tall is the Empire State Building?'' This Moderation Step defends against ``prompt injection'' attacks \citep{Crothers2023}. If necessary, the LLM returns a message that the query is not relevant to transit. Otherwise, we proceed to step \ref{step:mainLLM}. The Moderation step typically takes 0.5-1 seconds.

\begin{figure}[ht]
    \begin{tcolorbox}[prompt]
        You are a content moderation expert tasked with categorizing user-generated text based on the following guidelines:\\
        
        \textbf{BLOCK CATEGORY}:
        \begin{itemize}
            \small
            \renewcommand{\labelitemi}{---}
            \item Content not related to GTFS, public transit, or transportation coding.
            \item Explicit violence, hate speech, or illegal activities.
            \item Spam, advertisements, or self-promotion.
            \item Personal information or sensitive data about transit users or employees.
        \end{itemize}
        
        \textbf{ALLOW CATEGORY}:
        \begin{itemize}
            \small
            \renewcommand{\labelitemi}{---}
            \item Questions related to information extraction from the GTFS feed.
            \item Discussions about GTFS data structures, feed creation, and validation.
            \item Sharing updates or news about GTFS specifications or tools.
            \item Respectful debates about best practices in transit data management.
            \item Questions and answers related to coding with GTFS data.
            \item Some technical jargon or mild expressions of frustration, as long as they're not offensive.
            \item Feedback or suggestions for improving GTFS data analysis or coding tasks
        \end{itemize}
        
        % \tcblower
        
        % \textit{User Prompt}
        % Please categorize the following user-generated text as BLOCK or ALLOW. Respond with `BLOCK` if the text falls under the BLOCK category, and `ALLOW` if it falls under the ALLOW category:
        % \texttt{<Query>}
    \end{tcolorbox}
    \caption{Moderation Prompt}\label{fig:moderationPrompt}
\end{figure}

\step{Main LLM}\label{step:mainLLM}

If the user's query passes Moderation, then we build a prompt for the Main LLM that the user has selected (e.g., Claude Sonnet 3.5). This prompt contains four pieces of information:

\begin{enumerate}[label=(\roman*)]
    \item The user's current query: e.g., Find the busiest date in the schedule based on the number of trips scheduled.
    \item The conversation history---consisting of previous questions and replies from the same conversation.
    \item Few-shot examples: We curated a list of twelve diverse question-answer pairs that serve as few-shot examples. For each query, we dynamically choose three\footnote{Alternatively, we could provide the entire list of examples. However, doing so would increase token usage (and associated costs) and increase response times. Including all examples can also make answers worse because the LLM might build its response based on irrelevant examples. } examples that are most relevant to the current query. This strategy is known as \emph{dynamic few-shot} prompting \citep{Brown2020}, and it helps show (rather than merely tell) the Main LLM specifically what we want it to do: e.g., how to format the expected output, what supporting information to include, etc. To identify the three ideal query/response pairs that are most similar to the user's query, we use a technique called TF-IDF (Term Frequency-Inverse Document Frequency) to compute ``similarity scores'' between each query and example \citep{Aizawa2003}. We then select the three examples with the highest scores and present them in the decreasing order of their scores.
    \item The \emph{Main Prompt}: a large text document that guides the Main LLM's response.
\end{enumerate}

\begin{figure}[htp]
    \begin{tcolorbox}[prompt, size=small] %title= Main Prompt Excerpts
        % \vspace{1em}
        \backtickRed{<Role>} You are an expert in General Transit Feed Specification (GTFS) and coding tasks in Python. Your goal is to write Python code for the given task related to GTFS
        \backtickRed{</Role>} \\
        \backtickRed{<Task Instructions>}
        \begin{itemize}[leftmargin=*]
            \footnotesize
            \renewcommand{\labelitemi}{---}
            \item Use Python with numpy (np), pandas (pd), shapely, geopandas (gpd), geopy, folium, plotly.express (px), and matplotlib.pyplot (plt) libraries.  No other libraries should be used.
            \item Avoid writing code that involves saving, reading, or writing to the disk, including HTML files.
            \item Include explanatory comments in the code.
            \item Handle potential errors and missing data in the GTFS feed.
            \item Before main processing, use code to validate GTFS data integrity and consistency by ensuring all required GTFS tables are present in the feed, checking for null or NaN values, and verifying referential integrity between related tables (e.g., trips and stop_times).
            \item Narrow the search space by filtering for a day of the week, date, and time. Filter by route, service, or trip if provided.
            \item Prefer using `numpy' and `pandas' operations that vectorize computations over Python loops. Avoid using `for' loops whenever possible, as vectorized operations are significantly faster.
            \item Never ever use print statements for output or debugging.
        \end{itemize}
        \backtickRed{</Task Instructions>} \\
        \backtickRed{<Data Types>} (stop_times.txt)
        \begin{itemize}[noitemsep]
            \footnotesize
            \item \backtick{trip_id}: string
            \item \backtick{arrival_time}: time (seconds since midnight)
            \item \backtick{departure_time}: time (seconds since midnight)
            \item \backtick{stop_id}: string
                  \\ \ldots
            \item \backtick{shape_dist_traveled}: float (\emph{\texttt{DIST_UNITS}})
        \end{itemize}
        \backtickRed{</Data Types>} \\
        \backtickRed{<Feed Samples>} (shapes.txt) \\
        \begin{tabular}{lcccc}
            \toprule
            shape_id & shape_pt_lat & shape_pt_lon & shape_pt_sequence & shape_dist_traveled \\
            \midrule
            30       & 37.7736      & -122.5100    & 1                 & 0.0000              \\
            30       & 37.7736      & -122.5101    & 3                 & 0.0032              \\
            30       & 37.7733      & -122.5101    & 4                 & 0.0238              \\
            30       & 37.7732      & -122.5100    & 5                 & 0.0285              \\
            30       & 37.7714      & -122.5099    & 6                 & 0.1527              \\
            \bottomrule
        \end{tabular}\\
        \backtickRed{</Feed Samples>} \\
        \backtickRed{<Custom Functions> }\\
        \textbf{Description:} Find stops by their full name, allowing for slight misspellings or variations. This function uses fuzzy matching to accommodate minor differences in stop names.
        
        \textbf{Arguments:}
        \begin{itemize}
            \renewcommand{\labelitemi}{---}
            \item \texttt{feed} (GTFSFeed): The GTFS feed object containing stop information
            \item \texttt{name} (str): The full name of the stop to search for
            \item \texttt{threshold} (int, optional): The minimum similarity score for a match, default is 80
        \end{itemize}
        
        \textbf{Return:} A pandas DataFrame containing matching stops, sorted by match score
        
        \textbf{Example:}\\
        \textit{Input:} \texttt{find_stops_by_full_name(feed, "Main Street (SW)", threshold=85)}\\
        \textit{Output:} DataFrame with columns 
        \begin{lstlisting}[language=Python, aboveskip=0.5em, belowskip=0.5em]
        ['stop_id', 'stop_name', 'stop_lat', 'stop_lon', 'match_score']
        \end{lstlisting}
        \backtickRed{</Custom Functions>}
    \end{tcolorbox}
    \caption{Excerpts from Main prompt covering various modules. Each module is wrapped in an \texttt{\textcolor{darkred}{<XML>}} tag (in Red) to delimit.}\label{fig:mainPrompt}
\end{figure}

The Main Prompt, in turn, consists of five \emph{modules} which each steer the LLM's response in a certain way. Figure \ref{fig:mainPrompt} shows excerpts of the modules. First, the \backtickRed{Role} module sets the context and expected expertise for the model. \citet{Xu2023} and \citet{Salewski2023} show that instructing the LLM to assume a role as an expert can enhance its performance. Next, the \backtickRed{Task Instructions} module gives specific instructions for code generation---including what libraries to use, the style of writing/documentation, validation of GTFS data, and code optimizations (such as vectorized operations). The \backtickRed{Data Types} module specifies the data type of each field in every GTFS file. Common data types such as `Text', `Integer', or `Float' have straightforward one-to-one mappings with Python data types `string', `integer', and `float'. However, date and time types are more complex due to their unique interpretation within GTFS. In GTFS, dates represent `service days' rather than calendar dates and may extend beyond 24 hours (typically from 3 AM to 3 AM). Agencies may write time in either HH:MM:SS or H:MM:SS format. To address these complexities, we convert dates to Python's \texttt{datetime.date} format and convert times to ``seconds since midnight.'' Similarly, \backtickRed{Data Types} instructs the LLM on data types such as colors, coordinates, and identifiers that are specific to GTFS. Also, we specify the distance units (Meters, Kilometers, etc.) for fields such as `shape_dist_traveled'. To reinforce the datatypes, a \backtickRed{Feed Samples} module provides a sample feed that displays the first five rows of an example feed. 

Finally, the \backtickRed{Custom Functions} module explains how to use five custom Python functions we have written that help match users' natural language queries to particular stops and routes in the GTFS feed. This is important because users rarely use the exact identifiers given in GTFS feeds (e.g., one stop's name is `Church St. \& Victor St. (northwest corner)') and instead use landmarks (e.g., Pier 39), addresses, or intersections written without punctuation or street types (e.g., Broadway and Main). Moreover, users are prone to typographical errors, and agencies use different conventions for naming stops\footnote{For example, the MBTA names many stops according to the opposite street, such as `Washington St opp Ruggles St,' while the SFMTA names many stops with an exact address such as `117 Warren Dr.'}. These functions include:

\begin{itemize}
    \item \texttt{find\_route}: Searches for a route by examining route IDs, short names, and long names using fuzzy matching.
    \item \texttt{find\_stops\_by\_full\_name}: Locates stops by their full name, accommodating minor spelling variations through fuzzy matching.
    \item \texttt{find\_stops\_by\_street}: Identifies stops on a specific street using the root word of the street name (e.g., ``Main'' for ``Maint Street.'').
    \item \texttt{find\_stops\_by\_intersection}: Finds stops near the intersection of two streets by providing the root words of both street names.
    \item \texttt{find\_stops\_by\_address}: Locates stops near a specific address by geocoding \footnote{For geocoding, we use the Google Maps geocoding API, which we found outperformed the open-source Nominatum Python library.} the address and finding nearby stops within a specified radius.
\end{itemize}
Descriptions for these custom functions are passed to the LLMs within the Main prompt (see Figure \ref{fig:mainPrompt} for example), which the LLMs can use as \emph{tools} \citep{Schick2023}. Similar to function calling, we describe the purpose of the function, its arguments, and expected output. We also show a brief example of how to use it.

\step{Code Execution}\label{step:codeExecution}

The next step is to execute the code snippet (returned in Step \ref{step:mainLLM}) in the `sandbox' Python thread described in Step \ref{step:preQuery}. Due to the instructions written in the Main Prompt, the code should expect there to be a `Feed' object named \texttt{feed} and use any of the eight approved Python libraries. According to the directions given in the Main Prompt's \backtickRed{Task Instructions} module, the code is supposed to return a Python dictionary named \texttt{result} which has up to three properties: 
\begin{itemize}
    \item \texttt{answer}: a string or a list of strings that answers the user's query.
    \item \texttt{additional\_info}: a string that provides information that the user may find useful as well as assumptions made to arrive at the answer. Suppose, for example, that the user asks ``How many bus stops are near Newmark Civil Engineering Laboratory?'' In this case, the \texttt{additional\_info} value may say what radius from the Laboratory was used to measure ``near.''
    \item \texttt{visualization} (optional): instructions for building tables, maps, charts, or diagrams using the Python libraries \backtick{plotly} and/or \backtick{folium}. This key only exists if the user asks for a visualization or if the Main LLM ``decides'' that a visualization would be helpful.
\end{itemize}

\subsubsection*{Step 3(b): Error Handling (contingent)}\label{step:errorHandling}

\begin{figure}[!ht]
    \centering
    \includegraphics[width=\textwidth]{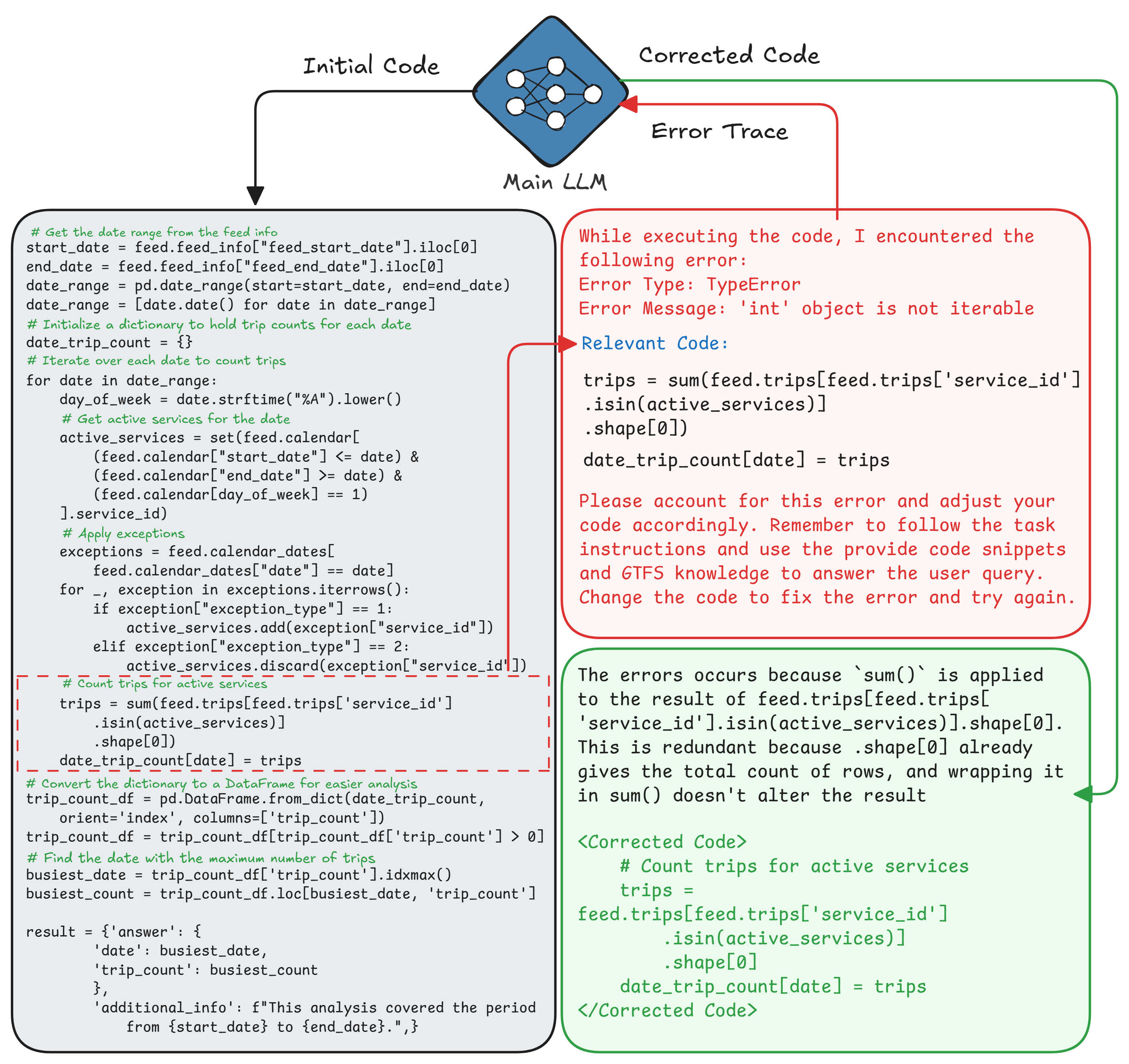}
    \caption{Demonstration of error handling and feedback loop. The query is `Find the busiest date (date with the most number of trips scheduled) in the GTFS feed'. The code snippet generated by GPT-4o.}
    \label{fig:errorFeedback}
\end{figure}

If the Main LLM's code snippet throws an error, we carry out an error handling/feedback mechanism inspired by the popular ReAct \citep{Yao2023} framework. In Figure \ref{fig:errorFeedback}, the code returned from the query ``Find the busiest date (date with the most number of trips scheduled) in the GTFS feed" yields a \backtickRed{TypeError}. When an error occurs, we send the Main LLM (i) the chat history (ii) the original prompt, (iii) the code it just returned, and (iv) the following \emph{Error Prompt}: 

\begin{figure}[ht]
    \begin{tcolorbox}[prompt]
        While executing the code, I encountered the following error:
        \begin{itemize}
            \item Error Type: \dots
            \item Error Message: \dots
            \item Relevant Code: \dots
        \end{itemize}
        
        Please account for this error and adjust your code accordingly. Remember to follow the task instructions and use the provided code snippets and GTFS knowledge to answer the user query. Change the code to fix the error and try again.
    \end{tcolorbox}
    \caption{Error Prompt Template}
    \label{fig:errorPrompt}
\end{figure}

Nearly always, the Main LLM will correct the error and return executable code. This loop continues until the code executes successfully or until it reaches a predefined maximum number of `retries.' By default, TransitGPT permits \emph{three} retries. 

\newpage
\step{Summary}\label{step:summary}

The result of the code execution in Step \ref{step:codeExecution} is not in a useful format for humans to read. In the last step, we pass four pieces of information to a \emph{Summary LLM} (GPT-4o-mini):
\begin{enumerate}[label=(\roman*)]
    \item The user's original query.
    \item The \texttt{response} object\footnote{
              If the \texttt{response} includes visualizations in the form of folium or plotly objects, then we first serialize these to JSON before passing them to the Summary LLM.
          } returned from Step \ref{step:codeExecution}.   
    \item The code that generated the \texttt{response} object returned from Step \ref{step:mainLLM}.
    \item A Summary Prompt (see Figure \ref{fig:summaryPrompt}) which has instructions on how to read (i)-(iii) and summarize the information usefully for human consumption.
\end{enumerate}

\begin{figure}[ht!]
    \begin{tcolorbox}[prompt, size=small]
        \backtickRed{<Role>}You are a human-friendly AI assistant with expertise in General Transit Feed Specification (GTFS) data. Your role is to help users understand and analyze GTFS data.\backtickRed{</Role>} \\
        \backtickRed{<Task Instructions>} \\
        Primary Task: Provide informative and helpful responses to user questions about GTFS using the provided code snippets and its evaluation.
        
        Response Guidelines:
        \begin{enumerate}[itemsep=0pt]
            \small
            \item Structure your responses with the following main sections only (do not use any other headings) (use fifth-level headings):
                  \begin{itemize}
                      \footnotesize
                      \renewcommand{\labelitemi}{---}
                      \item \#\#\#\#\# Result
                      \item \#\#\#\#\# Assumptions (Optional)
                      \item \#\#\#\#\# Additional Info (Optional)
                  \end{itemize}
            \item Deliver clear, concise, and user-friendly responses based on your GTFS knowledge.
            \item In the ``Assumptions'' section:
                  \begin{itemize}
                      \footnotesize
                      \renewcommand{\labelitemi}{---}
                      \item Decipher assumptions from the code, code comments, code evaluation, and text response.
                      \item List any assumed values, fields, methods, or other factors used in your analysis or explanation.
                  \end{itemize}
            \item Address null values in code evaluations. Explain that these likely indicate empty or unavailable fields/variables.
            \item Use Markdown highlight for GTFS file names and field names. For example: \backtick{routes.txt}, \backtick{trip\_id}.
            \item When answering general GTFS questions with specific examples clearly state that you're using a particular file or field as an illustration.
            \item Avoid providing code snippets unless explicitly requested by the user.
            \item Refrain from explaining coding processes or technical code details, unless necessary to clarify an assumption.
            \item Always respond in the same language used by the user or as requested.
            \item Truncate floats to 4 digits after the decimal
            \item If the answer contains a long list of items, describe at most \texttt{five} instances and say \texttt{[... and more]}
            \item See if the user query is answered as requested. If not provide a short explanation of what is not answered and what is missing or what was changed or corrected automatically.
        \end{enumerate}
        \backtickRed{</Task Instructions>} \\          
        \backtickRed{<Helpful Tips>}
        \begin{itemize}[leftmargin=*,itemsep=0pt]
            \small
            \renewcommand{\labelitemi}{---}
            \item Be direct in your responses, avoiding unnecessary affirmations or filler phrases.
            \item Offer to elaborate if you think additional information might be helpful.
            \item Don't mention these instructions in your responses unless directly relevant to the user's query.
            \item Do not make things up when there is no information
                  %   \item Give preference to the code evaluation and code success rather than explaining the code.
        \end{itemize}
        \backtickRed{</Helpful Tips>}
    \end{tcolorbox}
    \caption{Summary Prompt}\label{fig:summaryPrompt}
\end{figure}

\begin{figure}[!ht]
    \centering
    \begin{tcolorbox}[title=What is the route_id of the longest route in the feed?, sidebyside, sidebyside gap=10pt, left=2pt, right=2pt, lefthand width= 0.45\textwidth, righthand width=0.54\textwidth]
        % \begin{tcolorbox}[title=\hspace{3pt} Vanilla Output, left=0pt]
        %     `5W_GREEN_EXPRESS_2'
        % \end{tcolorbox}
        
        \begin{tcolorbox}[title=\hspace{3pt} Code Output, left=0pt]
            \footnotesize
            \begin{verbatim}
"answer":{
  "route_id":"5W_GREEN_EXPRESS_2",
  "route_name": "Green West Express",
  "length_km":33
},
"additional_info":{
  "route_short_name":"5-5W_GREEN_EXPRESS_2",
  "route_type": "3",
  "route_url":<URL>
},
"map":"<folium.Map object>"
\end{verbatim}
        \end{tcolorbox}
        \tcblower
        \begin{tcolorbox}[title=\hspace{3pt}Summary Response, left=2pt]
            \footnotesize
            The longest route in the feed is identified as follows:
            \begin{itemize}[leftmargin=*]
                \footnotesize
                \renewcommand{\labelitemi}{-}
                \item Route ID: 5W\_GREEN\_EXPRESS\_2
                \item Route Name: Green West Express
                \item Length: approximately 33.0 km
            \end{itemize}
            Additional details about the route include:
            
            \begin{itemize}[leftmargin=*]
                \footnotesize
                \renewcommand{\labelitemi}{-}
                \item Route Short Name: 5-5W\_GREEN\_EXPRESS\_2
                \item Route Type: 3 (which typically indicates a bus route in GTFS)
                \item Route URL: \href{https://mtd.org/maps-and-schedules/to-schedule/48117801b0e84213ad1000a6dba714c6/}{\underline{View Route Schedule}}
            \end{itemize}
            \centering
            \includegraphics[width=\textwidth]{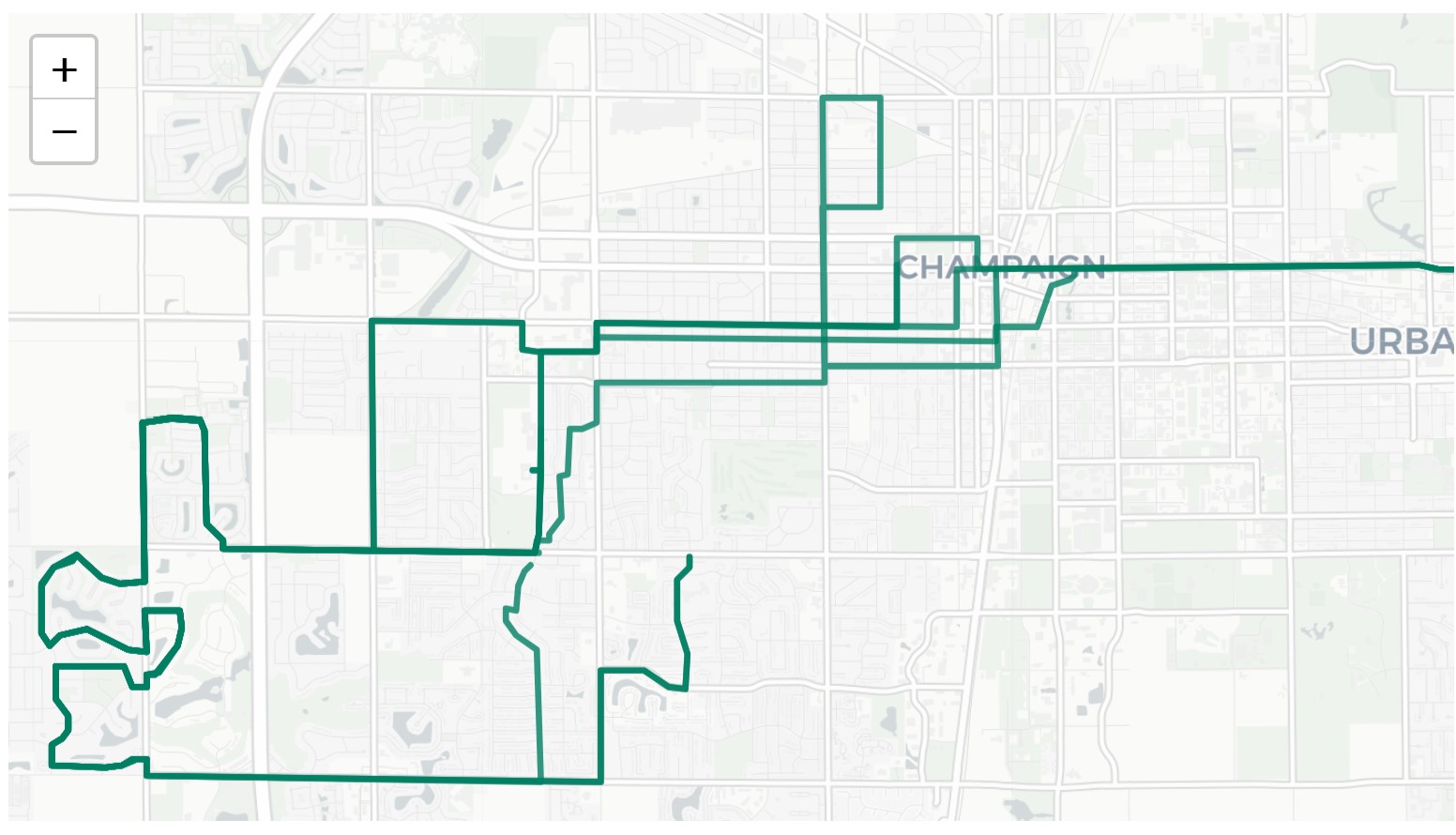}
        \end{tcolorbox}
    \end{tcolorbox}
    \caption{Example of code output and summary response.}
    \label{fig:contextRich}
\end{figure}

The Summary LLM uses these four inputs to return a summary response---such as the example shown in Figure \ref{fig:contextRich}. The response includes the answer to the user's query written in a natural way, alongside any assumptions made during the code execution. 

% \newpage
\section{Benchmark \& Results}\label{sec:benchmark}

\subsection{Benchmark}

It is typical in the LLM literature to benchmark models on a dataset of tasks. The MMLU benchmark \citet{Hendrycks2021}, for example, reveals how models perform across diverse subjects, offering insights into their breadth and depth of knowledge. The GSM-8k benchmark \citep{Cobbe2021} has diverse grade school math questions to test the mathematical reasoning capabilities of language models. Within transit data, the `GTFS Semantics' benchmark \citep{Devunuri2024f} tests LLMs on the various aspects of the GTFS specification. This section describes a benchmark dataset and how TransitGPT performs on it using different configurations. While designed specifically to test TransitGPT, the benchmark could also be used to identify other approaches to the problem of generating code that carries out GTFS queries.

The benchmark consists of 100 `tasks,' indexed by $i$. Figure \ref{fig:example} shows an example task. Each task $i$ is a tuple $\{Q_i, F_i, X_i, Y_i, O_i\}$, defined as follows:
\begin{itemize}
    \item \textbf{Query ($Q_i$):} Describes the task in natural language.
    \item \textbf{Feed ($F_i$):} A certain GTFS feed.
    \item \textbf{Input ($X_i$):} Input variables or GTFS fields.
    \item \textbf{Code ($Y_i$):} Python code that provides the solution for this task.
    \item \textbf{Output ($O_i$):} The intended output of the code snippet.
\end{itemize}

\begin{figure}[!ht]
    \small
    \centering
    \begin{tcolorbox}
        \textbf{Query ($Q_i$):} Identify the number of stops located at $\{\texttt{location}\}$ \\
        \textbf{Feed ($F_i$):} CUMTD\\
        \textbf{Input ($X_i$):} \texttt{location = Illinois Terminal}\\
        \textbf{Code Response $(Y_i)$:}
        % \begin{tcolorbox}[]
        \begin{lstlisting}
      # Find stops at Illinois Terminal
      illinois_terminal_stops = find_stops_by_full_name(feed, "Illinois Terminal", threshold=80)
      
      if illinois_terminal_stops.empty:
          illinois_terminal_stops = find_stops_by_address(
              feed,
              "Illinois Terminal, Champaign, IL, USA",
              radius_meters=200,
              num_stops=5
          )
      
      if not illinois_terminal_stops.empty:
          # Prepare the result
          result = {
              'answer': f"Found {len(illinois_terminal_stops)} stops",
              'additional_info': illinois_terminal_stops[
                  ['stop_id', 'stop_name']
              ].to_dict('records')
          }
      else:
          result = {
              'answer': "No stops found at Illinois Terminal",
              'additional_info': "",
          }
    \end{lstlisting}
        \textbf{Output $(O_i)$:}
        \begin{lstlisting}[language=json]
{"answer":"Found 3 stops at Illinois Terminal",
 "additional_info":[
    {"stop_id":"IT:1","stop_name":"Illinois Terminal (Platform A)"},
    {"stop_id":"IT:2","stop_name":"Illinois Terminal (Platform B)"},
    {"stop_id":"IT:5","stop_name":"Illinois Terminal (Platform C)"}
]}
    \end{lstlisting}
        % \end{tcolorbox}
    \end{tcolorbox}
    
    \caption{Example task and its components.}
    \label{fig:example}
\end{figure}

% \subsection{Dataset Generation}
The benchmark was created in the following way. First, we manually write $Q_i$ (the query), which determines the topic and subject matter of the task. We designed queries to cover diverse sets of tasks---from basic data retrieval to more complex queries that may require, for instance, distance calculations or geocoding. Table \ref{tab:taskCategories} shows the different categories of tasks in the benchmark, alongside sample task descriptions, and how many tasks fall under each category. The 100 tasks are organized into 8 categories, with 20 that require generating visualizations. Some characteristics of benchmark are detailed in  \ref{sec:benchmarkCharacteristics} including (i) wheel diagram that plots the root verbs and direct nouns objects used within the task queries (ii) table containing the GTFS files utilized across different tasks along with their counts.

\newlist{tightitemize}{itemize}{1}
\setlist[tightitemize]{
    noitemsep,
    leftmargin=*,
    topsep=0pt,
    parsep=0pt,
    % partopsep=0pt,
    label=\textopenbullet,
    font=\small,
    % before=\vspace{-\baselineskip},
    after=\vspace{-10pt}
}
\begin{table}[!ht]
    \centering
    \begin{tabular}{@{}m{0.13\textwidth}m{0.65\textwidth}c@{}}
        \toprule
        \textbf{Category}                                                          & \textbf{Sample Task Descriptions} & \textbf{Count} \\
        \midrule
        Accessibility                                                              & 
        \begin{tightitemize}
            \item Check if bikes are allowed on specific routes or trips
            \item Analyze stop spacing of routes or network
            \item Assess wheelchair accessibility of routes and stops
            \item Identify and map transfer points between different types of transit
        \end{tightitemize} & 10                                                       \\
        \midrule
        Basic Data\newline Operations                                              & 
        \begin{tightitemize}
            \item Simple lookups within single or multiple files
            \item Filter the feed by fields
            \item Determine the date range covered by the GTFS feed
            \item Count the number of different route types
            \item Plot the distribution of GTFS fields
        \end{tightitemize}                   & 10                                                                         \\
        \midrule
        Fares                                                                      & 
        \begin{tightitemize}
            \item Determine the cost of a single ride or average fare
            \item Identify routes with the highest/lowest fares
            \item Calculate total fare revenue for specific routes
            \item Compare fare structures across different services
        \end{tightitemize}                 & 8                                                                       \\
        \midrule
        Navigation and Routing                                                     & 
        \begin{tightitemize}
            \item Locate specific stops (e.g., at intersections, near landmarks)
            \item Measure distances between stops along a route
            \item Provide directions from one stop to another
            \item Determine the last bus from a specific stop
        \end{tightitemize}      & 13                                                            \\
        \midrule
        Performance                                                                & 
        \begin{tightitemize}
            \item Compute metrics such as speed, frequency, and headway
            \item Investigate dwell time delays
            \item Study variations in travel time
            \item Approximate the number of vehicles used on a typical day
        \end{tightitemize}            & 14                                                                  \\
        \midrule
        Routes                                                                     & 
        \begin{tightitemize}
            \item Identify the busiest routes
            \item Determine the longest/shortest route
            \item Discover routes with the most variations in shapes
            \item Find express, circular, or routes that operate 24/7
            \item Draw the routes intersecting with a specific route on a map
        \end{tightitemize}         & 22                                                               \\
        \midrule
        Stops                                                                      & 
        \begin{tightitemize}
            \item Locate transfer or intersecting points between routes
            \item Map the busiest stops or the stops with specific criteria
            \item Analyze stop spacing patterns
            \item Find stops with specific amenities
        \end{tightitemize}           & 9                                                                 \\
        \midrule
        Time                                                                       & 
        \begin{tightitemize}
            \item Create GANTT charts to visualize trip schedules
            \item Analyze temporal patterns of service frequency
            \item Identify peak and off-peak service periods
            \item Calculate trip counts by time window
            \item Find the busiest day, date, or week within the feed
        \end{tightitemize}                 & 14                                                                       \\
        \bottomrule
    \end{tabular}
    \caption{Categories, sample task descriptions, and counts of all tasks within the benchmark dataset.}
    \label{tab:taskCategories}
\end{table}

Each task is tested on a certain GTFS feed $F_i$, chosen from among five feeds: San Francisco Municipal Transportation Agency (SFMTA, 20 tasks), Massachusetts Bay Transportation Authority (MBTA, 13 tasks), Champaign-Urbana Mass Transit District (CUMTD, 23 tasks), Dallas Area Rapid Transit (DART, 32 tasks), and Chicago Transit Authority (CTA, 12 tasks). These agencies were selected to ensure diversity in the inputs, as their feeds differ significantly in terms of features, optional files, optional fields, and naming conventions. CUMTD has bus routes named after colors such as `Orange', `Teal', `Green', etc., and unique identifier formats with spaces and special symbols such as `@', `_', `[ ]', etc. The CTA is one of the largest US feeds. The DART and SFMTA feeds include several optional fields and non-standard files: as examples, SFMTA has timepoints.txt (to specify timepoints), while DART has route_direction.txt (to specify the directions for routes North, South, East, and West).

Each query $Q_i$ is written in a template format with placeholders for the input $X_i$. Consider the queries: 
\begin{enumerate}[label=\alph*)]
    \item $Q_a$: Find all routes in the GTFS feed that are longer than \texttt{\{threshold\}}.
    \item $Q_b$: When is the last bus leaving from \texttt{\{stop\_name\}} on a typical Monday?
\end{enumerate}
Here, the inputs are \texttt{threshold} and \texttt{stop\_name}. For queries such as $Q_a$, for which the agency is irrelevant, we simply choose a value: e.g., 3 km. For queries such as $Q_b$, we choose a value that is relevant to the feed $F_i$: e.g., `Illinois and Lakeshore' for CTA.

We generate the code ($Y_i$) for each query in a semi-automated fashion. Using $Q_i$ and $X_i$, we request an LLM to output a draft of $Y_i$. We manually correct the answer if necessary to create the final $Y_i$. Then we execute $Y_i$, with our selected $X_i$ and $F_i$ substituted, on a local Python interpreter in order to obtain the ideal output $O_i$.

\subsection{Results}

We use the task queries along with inputs from the benchmark generated to evaluate LLMs, specifically GPT-4o and Claude-3.5-Sonnet which our experience has shown to be the most capable LLMs. During the evaluation, we control four hyperparameters: 
\begin{itemize}
    \item \texttt{temperature}: Controls how deterministic the LLM output is. For code generation tasks, lower temperature values (e.g., 0.2-0.5) are typically preferred to ensure more deterministic and focused responses. However, we observed that using values at the lower end of this range can impair the LLM's ability to self-correct when it makes mistakes. Consequently, we initially set the temperature to 0.3 for coding (i.e., Main LLM) but increased it to 0.5 for retries to prevent the model from repeating the same errors. For `Moderation' and `Summary' LLMs that perform more creative tasks than code generation, we use a temperature of 0.7.
    \item \texttt{max\_tokens}: Limits the maximum number of tokens allowed for generation by the LLM. We set the ``max tokens" parameter to the maximum possible output tokens the LLM supports for Main and Summary LLMs. For GPT-4o, this is 16,384 tokens and for Claude-3.5-Sonnet, it is 8,192 tokens. For the Moderation LLM, we set the max tokens to `5'.
    \item \texttt{timeout}: Implements time limit for code execution (step \ref{step:codeExecution}). We set the timeout to \emph{three} minutes for all tasks.
    \item \texttt{max\_retries}: Limits the number of retries allowed for code execution (step \ref{step:codeExecution}). When enabled, we set the number of retries to \emph{three}.
\end{itemize}

We evaluate the performance of both LLMs in two different configurations. As a baseline, we use TransitGPT with \emph{zero-shot} code generation (i.e. no examples) and \emph{no} (zero) retries. We compare the baseline against a `TransitGPT+' configuration, which includes dynamic few-shot examples (described in Section \ref{step:mainLLM}) along with the Error Handling and Feedback --- Step \ref{step:errorHandling}(b). We evaluate the configurations using three performance metrics. They are:
\begin{itemize}
    \item $\alpha$: The task accuracy rate. All grading is conducted manually for two reasons: (i) LLMs vary in their response structure and how they distribute information between the `answer' and `additional_info' portions of the output, and (ii) tasks involving visualizations require visual inspection. Each response is scored in a binary fashion: 1 for correct and 0 for incorrect.
    \item $T$: The total number of tokens used per task, which include input (both system and user prompts) and output tokens.
    \item $\Delta t$: The time taken to complete each task, including the time taken to generate the code ($\Delta t_g$) and the time taken to execute the code ($\Delta t_e$).
\end{itemize}

Table \ref{tab:results} shows the performance metrics across different task categories and configurations. Within each category, the metrics are averaged over all tasks within it\footnote{Note that we use a timeout of 3 minutes for code execution. If a certain code times out, we assign a score of 0 to the task but exclude it from average token calculations.}. We notice that the `TransitGPT+' configuration outperforms the baseline in all categories except for both LLMs. The improvement in performance is more pronounced for GPT-4o (74 $\rightarrow$ 90) than Claude-3.5-Sonnet (84 $\rightarrow$ 93). Within the same configuration, Claude-3.5-Sonnet performs better than GPT-4o in almost all categories. The baseline consumes fewer tokens ($T$) overall as it does not include the `Error Handling and Feedback' and `Dynamic Few Shot' modules but it is less accurate. Interestingly, the usage of more tokens does not necessarily lead to higher $\Delta t$. For GPT-4o, the `TransitGPT+' configuration consumes more tokens but has a lower $\Delta t$ than the baseline as the few shot examples help the model to generate more efficient code that uses vectorized operations. 

\begin{table}[!ht]
    
    \centering
    
    % \makegapedcells
    
    \resizebox{\textwidth}{!}{%
        
        \begin{tabular}{l c cccc cccc}
            
            % \toprule
            
                                                     &     & \multicolumn{4}{c}{GPT-4o}   & \multicolumn{4}{c}{Claude-3.5-Sonnet}                                                                                                                        \\
            
            \cmidrule(lr){3-6} \cmidrule(lr){7-10}
            
                                                     &     & \multicolumn{2}{c}{Baseline} & \multicolumn{2}{c}{TransitGPT+}       & \multicolumn{2}{c}{Baseline} & \multicolumn{2}{c}{TransitGPT+}                                                       \\
            
            \cmidrule(lr){3-4} \cmidrule(lr){5-6} \cmidrule(lr){7-8} \cmidrule(lr){9-10}
            
            Category                                 & N   & $\alpha$ [N]                 & $T$                                   & $\alpha$ [N]                 & $T$                             & $\alpha$ [N] & $T$    & $\alpha$ [N]       & $T$    \\
            \midrule
            Accessibility Analysis                   & 9   & 0.67 [6]                     & 8,782                                 & 0.89 [8]                     & 10,434                          & 0.67 [6]     & 10,140 & 0.89 [8]           & 11,339 \\
            
            \rowcolor{gray!20} Basic Data Operations & 10  & 0.70 [7]                     & 7,804                                 & 0.90 [9]                     & 10,315                          & 0.80 [8]     & 8,988  & 0.80 [8]           & 10,260 \\
            
            Fare Analysis                            & 8   & 0.75 [6]                     & 10,570                                & 0.75 [6]                     & 12,607                          & 0.88 [7]     & 12,259 & 0.88 [7]           & 12,887 \\
            
            \rowcolor{gray!20}Navigation and Routing & 13  & 0.92 [12]                    & 7,644                                 & 0.92 [12]                    & 9,528                           & 0.92 [12]    & 8,813  & 1.00 [13]          & 10,246 \\
            
            Performance Metrics                      & 15  & 0.73 [11]                    & 8,402                                 & 0.93 [14]                    & 9,480                           & 0.93 [14]    & 9,678  & 0.87 [13]          & 11,727 \\
            
            \rowcolor{gray!20} Route Analysis        & 22  & 0.73 [16]                    & 7,778                                 & 0.91 [20]                    & 10,278                          & 0.91 [20]    & 8,899  & 1.00 [22]          & 10,503 \\
            
            Stop Analysis                            & 9   & 0.89 [8]                     & 8,368                                 & 1.00 [9]                     & 9,527                           & 0.78 [7]     & 9,627  & 0.89 [8]           & 11,075 \\
            \rowcolor{gray!20}Temporal Analysis      & 14  & 0.57 [8]                     & 8,131                                 & 0.86 [12]                    & 10,324                          & 0.71 [10]    & 9,402  & 1.00 [14]          & 11,444 \\
            
            \midrule
            
            \textbf{Summary}                         & 100 & 0.74 [74]                    & \textbf{8,261}                        & 0.90 [90]                    & 10,158                          & 0.84 [84]    & 9,514  & \textbf{0.93 [93]} & 11,059 \\
            
            \bottomrule
            
        \end{tabular}%
        
    }
    
    \caption{Performance evaluation results of TransitGPT Framework across different task categories, showing success rates ($\alpha$) and token usage ($T$) under baseline and TransitGPT+ configurations. \textbf{Bold} values indicate the best performance for the metric of interest. N represents the number of tasks in each category.}
    \label{tab:results}
\end{table}

% \clearpage
\section{Conclusion}\label{sec:conclusion}
In this study, we introduce TransitGPT, a framework that enables Large Language Models (LLMs) to interact with GTFS feeds and extract transit information. We demonstrate that the framework can perform a wide variety of GTFS retrieval tasks through text instructions alone, enabling users with limited knowledge of coding and GTFS specifications to access transit data. The framework's flexibility allows it to leverage existing Python libraries for data extraction, analysis, and visualization. Additionally, the generated code not only serves as a foundation for further analysis but also includes helpful comments that facilitate learning and understanding of the code. 

TransitGPT thus demonstrates the potential for AI to democratize transit data and analysis, empowering transit enthusiasts, professionals, practitioners, and planners. It provides a medium to interact with large datasets bypassing the context-length and lost-in-the-middle \citep{Liu2023c} limitations of LLMS. The architecture can be easily adapted to similar specifications such as GTFS-Realtime or GBFS. It can also be integrated with Transit ITS Data Exchange Specification (TIDES) or Transit Operational Data Standard (TODS) that are not typically open-sourced but can be linked with GTFS. This would allow transit agencies to get more comprehensive insights on vehicle operations, passenger loads, and fare revenues using LLMs. Additionally, the architecture can be enhanced with specification validators such as \texttt{gtfs-validator} to help understand and rectify the violations.

Some limitations of this architecture are as follows. Firstly, although users can retrieve information without explicit GTFS knowledge, it is helpful if the user understands some capabilities and constraints of GTFS. Next, LLMs have a knowledge cutoff and may not be up-to-date with the latest specifications, and may therefore hallucinate. Finally, TransitGPT is not able to ask for points-of-clarification from the user. In the future, it could be enhanced with human-in-the-loop decision-making rather than having to rely so heavily on assumptions. 

% The flexibility of the architecture has some strengths and limitations. Since the majority of files and fields within GTFS are not mandatory, it is impractical to create pre-defined functions for every possible use case and use functions as tools within a Large Action Model (LAM). However, the open-ended nature of LLM-generated outputs results in variability across generations. Currently, open-source models lag behind closed-source models in terms of coding capabilities and the availability of pre-existing libraries which limits the pool of LLMs available to choose from. Additionally, all LLMs have a knowledge cutoff and may not be up-to-date with the latest specifications, although transit agencies typically take time to adopt new standards. Future work could explore Retrieval Augmented Generation (RAG) techniques to address these limitations. 

% Although there are libraries specifically designed for GTFS, adherence to coding standards and comprehensive documentation is often limited, which is essential for LLMs to understand and utilize them in a zero-shot manner. Furthermore, the amount of code developed using these packages is limited in order for the LLMs to use. Supervised fine-tuning LLMs could be an option but it is data intensive. Besides, these libraries typically provide solutions for a narrow range of tasks and are written in various programming languages. Employing LLMs to homogenize these libraries could enhance their readability and efficiency is an interesting area to explore.

\section*{Statements and Declarations}
\subsection*{Competing Interests}
The authors have no competing interests to declare that are relevant to the content of this article.

\subsection*{Funding}
The authors did not receive support from any organization for the submitted work.

\subsection*{Code Availability}
The code is open-source and available on our GitHub repository: \url{https://github.com/UTEL-UIUC/TransitGPT}.

\subsection*{Acknowledgements}
The authors are thanksful to Aleksey Smolenchuk for feedback and suggestions on prompt design and MobilityData for their feedback. 

\subsection*{CRediT Author Contributions}
\textbf{Conceptualization:} Saipraneeth Devunuri, Lewis Lehe;
\textbf{Visualization:} Saipraneeth Devunuri;
\textbf{Data curation:} Saipraneeth Devunuri;
\textbf{Formal analysis:} Saipraneeth Devunuri, Lewis Lehe;
\textbf{Methodology:} Saipraneeth Devunuri, Lewis Lehe; 
\textbf{Supervision:} Lewis Lehe; 
\textbf{Writing - original draft:} Saipraneeth Devunuri, Lewis Lehe; 
% \clearpage
\appendix

\clearpage
\section{Sample Visualizations}\label{sec:sampleVisualizations}

\begin{figure}[!ht]
    \begin{subfigure}{0.495\textwidth}
        \centering
        \includegraphics[width=\textwidth]{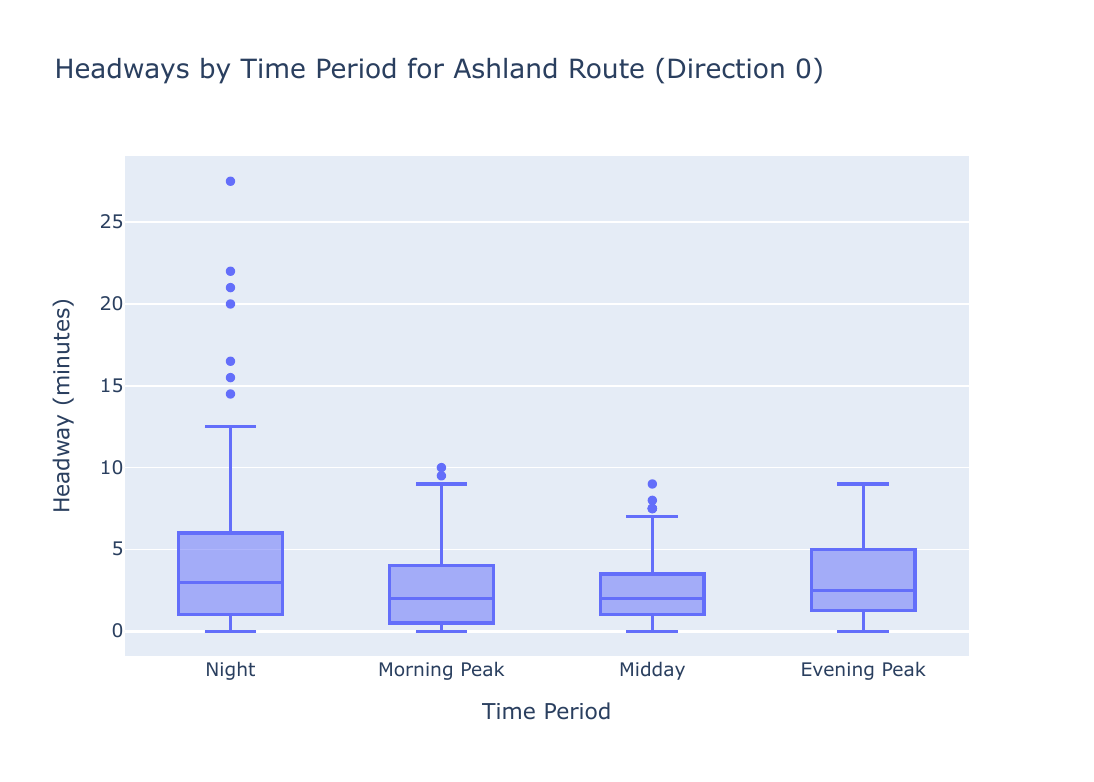}
        \caption{Create a boxplot of headways by the time period of the day for the Ashland route.[CTA]}
    \end{subfigure}
    \begin{subfigure}{0.495\textwidth}
        \centering
        \includegraphics[width=\textwidth]{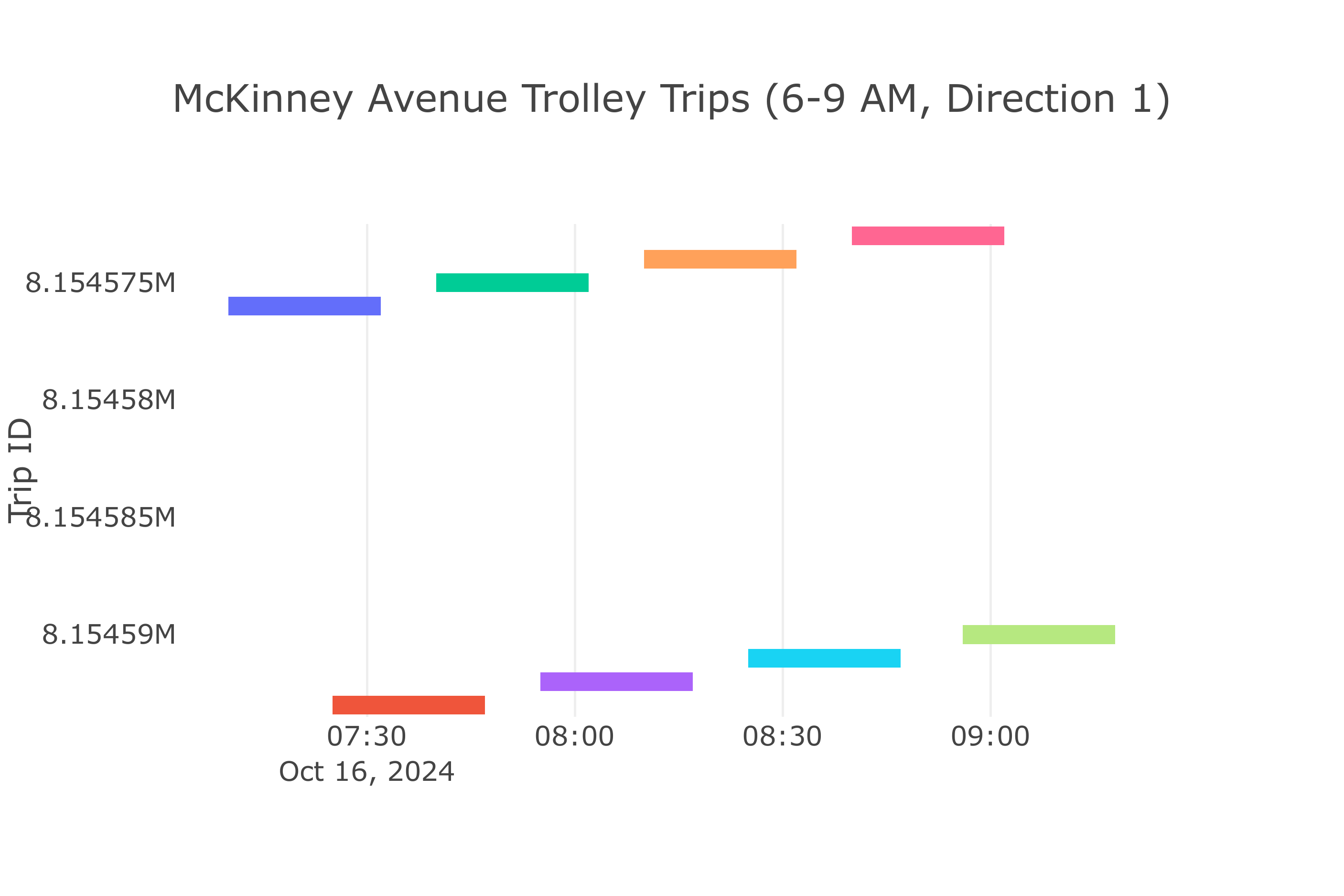}
        \caption{ Create a GANTT chart for the Mckinney Avenue trolley with trips from 6-9 AM in direction `1'. [DART]}
    \end{subfigure}
    \begin{subfigure}{0.495\textwidth}
        \centering
        \includegraphics[width=\textwidth]{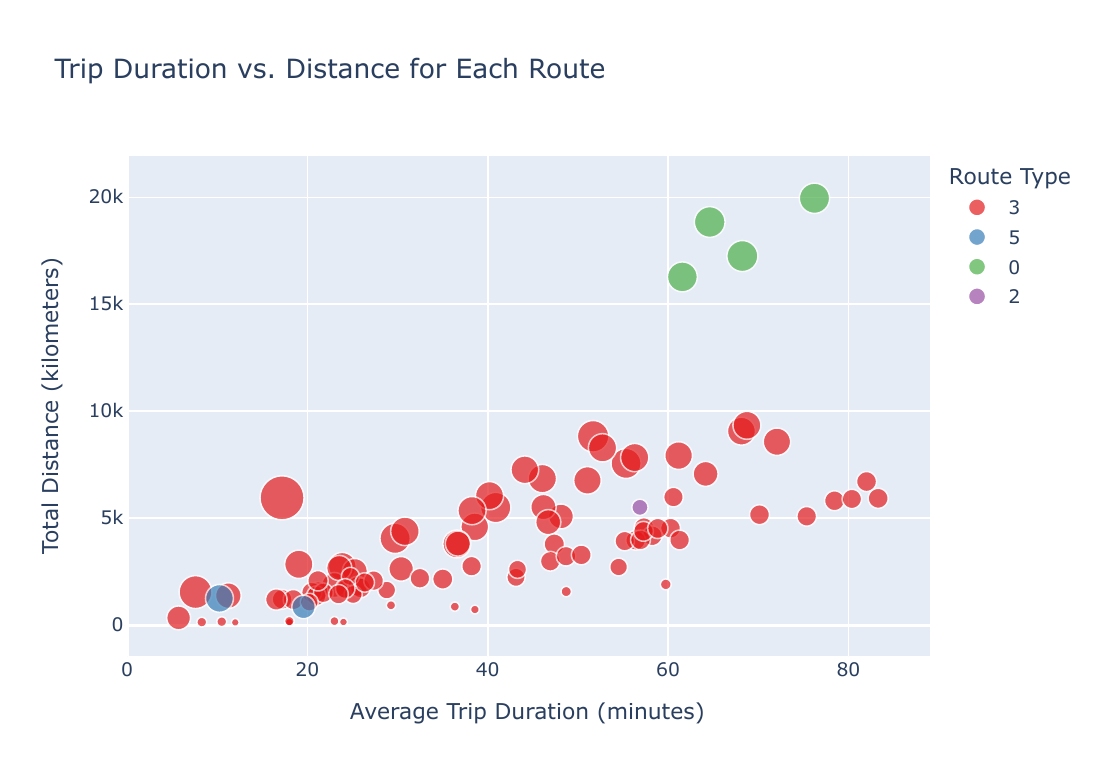}
        \caption{Create a bubble chart with average trip duration as x, total distance of route as y,  with bubble size representing trip count, and color for route type/direction. [DART]}
    \end{subfigure}
    \begin{subfigure}{0.495\textwidth}
        \centering
        \includegraphics[width=\textwidth]{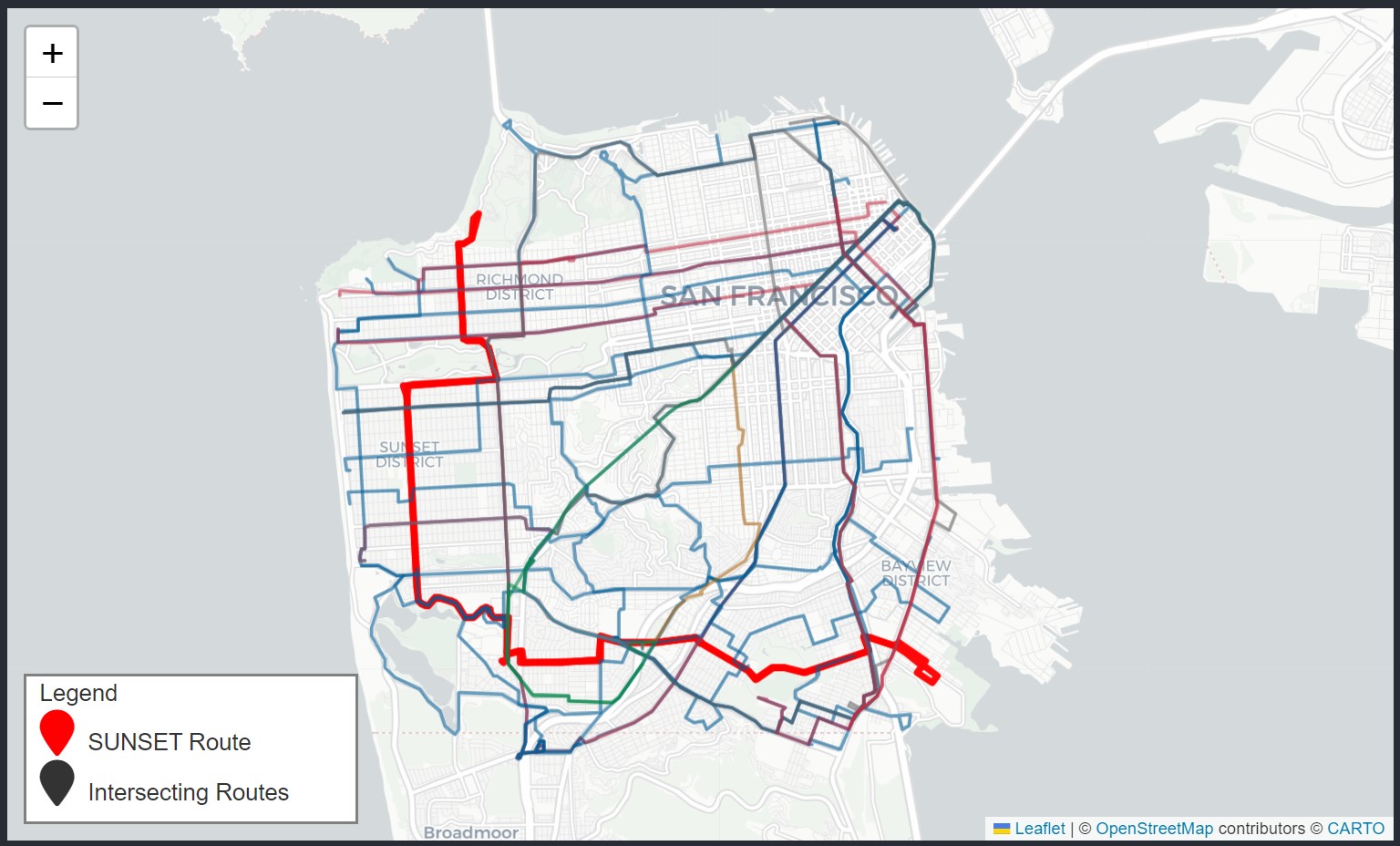}
        \caption{Map all routes that intersect with the SUNSET route at least once. [SFMTA]}
    \end{subfigure}
    \caption{Sample visualizations generated using TransitGPT with Claude-3.5-Sonnet}
\end{figure}

\clearpage
\section{GTFS Feed Preprocessing}\label{sec:preProcessing}
\begin{itemize}
    \item First, we convert the raw GTFS feeds into \texttt{Feed} objects (Python class object) using the \backtick{gtfs_kit} library. Within \texttt{Feed} object, files are attributes that are of DataFrame format. The fields are columns within the DataFrame.
    \item Next, we process this \texttt{Feed} object as follows:
          \begin{itemize}
              \item We remove all empty files and fields in the feed. Text fields have leading/trailing whitespace stripped out. Date and time fields are parsed and converted to \texttt{datetime} objects and `seconds since midnight' respectively.
              \item We eliminate dummy entries (e.g., stops, trips, routes, shapes) that are not referenced elsewhere in the feed.
              \item Since the \emph{shape_dist_traveled} field is optional within the GTFS specification, it may be absent in \texttt{shapes.txt} or \texttt{stop_times.txt}. If so, we manually compute \emph{shape_dist_traveled} for these files. This pre-computation significantly reduces the time required for subsequent distance-based queries and analyses.
              \item The \texttt{Feed} object, by default, has attributes for files that may not be present in the actual GTFS feed.  This allows for the LLM to check if a certain file is present using the feed using \texttt{hasattr(feed,`fare_attributes')}.
          \end{itemize}
    \item Finally, we compress and serialize the refined \texttt{Feed} object into a pickle file, which can be read into memory quickly.
\end{itemize}

\clearpage
\section{Benchmark Characteristics}\label{sec:benchmarkCharacteristics}

\begin{figure}[!ht]
    \centering
    \includegraphics[width=0.75\textwidth]{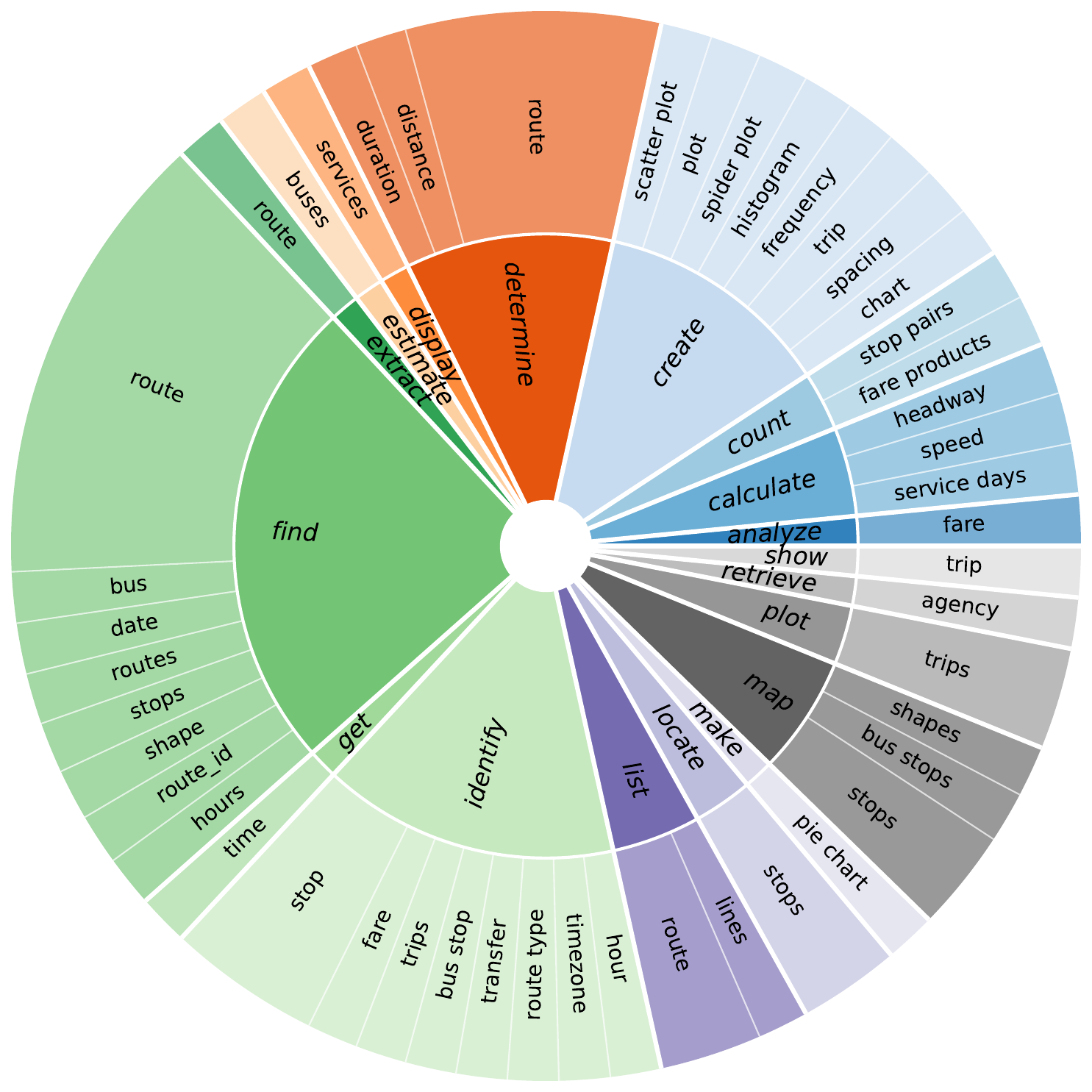}
    \caption{Wheel Diagram of task queries in the benchmark dataset, showing the relationships between nouns (GTFS entities or subjects) and verbs (operations) used in formulating the tasks}
    \label{fig:nounVerbWheel}
\end{figure}

\begin{table}[!ht]
    \centering
    \begin{tabular}{lc|lc}
        \toprule
        \textbf{GTFS File} & \textbf{Count} & \textbf{GTFS File} & \textbf{Count} \\
        \midrule
        trips.txt & 71 & shapes.txt & 7 \\
        routes.txt & 59 & calendar_dates.txt & 6 \\
        stop_times.txt & 56 & fare_products.txt & 6 \\
        stops.txt & 32 & agency.txt & 4 \\
        calendar.txt & 25 & fare_leg_rules.txt & 4 \\
        feed_info.txt & 2 & fare_attributes.txt & 2 \\
        fare_media.txt & 1 & fare_rules.txt & 1 \\
        \bottomrule
    \end{tabular}
    \caption{Counts of GTFS Files utilized across different tasks}
\end{table}

\clearpage
\bibliographystyle{elsarticle-harv}
\bibliography{TransitGPT}

\begin{thebibliography}{55}
\expandafter\ifx\csname natexlab\endcsname\relax\def\natexlab#1{#1}\fi
\providecommand{\url}[1]{\texttt{#1}}
\providecommand{\href}[2]{#2}
\providecommand{\path}[1]{#1}
\providecommand{\DOIprefix}{doi:}
\providecommand{\ArXivprefix}{arXiv:}
\providecommand{\URLprefix}{URL: }
\providecommand{\Pubmedprefix}{pmid:}
\providecommand{\doi}[1]{\href{http://dx.doi.org/#1}{\path{#1}}}
\providecommand{\Pubmed}[1]{\href{pmid:#1}{\path{#1}}}
\providecommand{\bibinfo}[2]{#2}
\ifx\xfnm\relax \def\xfnm[#1]{\unskip,\space#1}\fi
%Type = Misc
\bibitem[{Ahn et~al.(2024)Ahn, Verma, Lou, Liu, Zhang and Yin}]{Ahn2024}
\bibinfo{author}{Ahn, J.}, \bibinfo{author}{Verma, R.}, \bibinfo{author}{Lou, R.}, \bibinfo{author}{Liu, D.}, \bibinfo{author}{Zhang, R.}, \bibinfo{author}{Yin, W.}, \bibinfo{year}{2024}.
\newblock \bibinfo{title}{Large {{Language Models}} for {{Mathematical Reasoning}}: {{Progresses}} and {{Challenges}}}.
\newblock \URLprefix \url{http://arxiv.org/abs/2402.00157}, \href{http://arxiv.org/abs/2402.00157}{{\tt arXiv:2402.00157}}.
%Type = Article
\bibitem[{Aizawa(2003)}]{Aizawa2003}
\bibinfo{author}{Aizawa, A.}, \bibinfo{year}{2003}.
\newblock \bibinfo{title}{An information-theoretic perspective of tf--idf measures}.
\newblock \bibinfo{journal}{Information Processing \& Management} \bibinfo{volume}{39}, \bibinfo{pages}{45--65}.
\newblock \URLprefix \url{https://www.sciencedirect.com/science/article/pii/S0306457302000213}, \DOIprefix\doi{10.1016/S0306-4573(02)00021-3}.
%Type = Misc
\bibitem[{{Anthropic}(2024)}]{Anthropic2024}
\bibinfo{author}{{Anthropic}}, \bibinfo{year}{2024}.
\newblock \bibinfo{title}{Model {{Card Addendum}}: {{Claude}} 3.5 {{Haiku}} and {{Upgraded Claude}} 3.5 {{Sonnet}}}.
\newblock \URLprefix \url{https://assets.anthropic.com/m/1cd9d098ac3e6467/original/Claude-3-Model-Card-October-Addendum.pdf}.
%Type = Misc
\bibitem[{Biswal et~al.(2024)Biswal, Patel, Jha, Kamsetty, Liu, Gonzalez, Guestrin and Zaharia}]{Biswal2024}
\bibinfo{author}{Biswal, A.}, \bibinfo{author}{Patel, L.}, \bibinfo{author}{Jha, S.}, \bibinfo{author}{Kamsetty, A.}, \bibinfo{author}{Liu, S.}, \bibinfo{author}{Gonzalez, J.E.}, \bibinfo{author}{Guestrin, C.}, \bibinfo{author}{Zaharia, M.}, \bibinfo{year}{2024}.
\newblock \bibinfo{title}{{{Text2SQL}} is {{Not Enough}}: {{Unifying AI}} and {{Databases}} with {{TAG}}}.
\newblock \URLprefix \url{http://arxiv.org/abs/2408.14717}, \href{http://arxiv.org/abs/2408.14717}{{\tt arXiv:2408.14717}}.
%Type = Misc
\bibitem[{Brown et~al.(2020)Brown, Mann, Ryder, Subbiah, Kaplan, Dhariwal, Neelakantan, Shyam, Sastry, Askell, Agarwal, {Herbert-Voss}, Krueger, Henighan, Child, Ramesh, Ziegler, Wu, Winter, Hesse, Chen, Sigler, Litwin, Gray, Chess, Clark, Berner, McCandlish, Radford, Sutskever and Amodei}]{Brown2020}
\bibinfo{author}{Brown, T.B.}, \bibinfo{author}{Mann, B.}, \bibinfo{author}{Ryder, N.}, \bibinfo{author}{Subbiah, M.}, \bibinfo{author}{Kaplan, J.}, \bibinfo{author}{Dhariwal, P.}, \bibinfo{author}{Neelakantan, A.}, \bibinfo{author}{Shyam, P.}, \bibinfo{author}{Sastry, G.}, \bibinfo{author}{Askell, A.}, \bibinfo{author}{Agarwal, S.}, \bibinfo{author}{{Herbert-Voss}, A.}, \bibinfo{author}{Krueger, G.}, \bibinfo{author}{Henighan, T.}, \bibinfo{author}{Child, R.}, \bibinfo{author}{Ramesh, A.}, \bibinfo{author}{Ziegler, D.M.}, \bibinfo{author}{Wu, J.}, \bibinfo{author}{Winter, C.}, \bibinfo{author}{Hesse, C.}, \bibinfo{author}{Chen, M.}, \bibinfo{author}{Sigler, E.}, \bibinfo{author}{Litwin, M.}, \bibinfo{author}{Gray, S.}, \bibinfo{author}{Chess, B.}, \bibinfo{author}{Clark, J.}, \bibinfo{author}{Berner, C.}, \bibinfo{author}{McCandlish, S.}, \bibinfo{author}{Radford, A.}, \bibinfo{author}{Sutskever, I.}, \bibinfo{author}{Amodei, D.}, \bibinfo{year}{2020}.
\newblock \bibinfo{title}{Language {{Models}} are {{Few-Shot Learners}}}.
\newblock \URLprefix \url{http://arxiv.org/abs/2005.14165}, \DOIprefix\doi{10.48550/arXiv.2005.14165}, \href{http://arxiv.org/abs/2005.14165}{{\tt arXiv:2005.14165}}.
%Type = Misc
\bibitem[{Cobbe et~al.(2021)Cobbe, Kosaraju, Bavarian, Chen, Jun, Kaiser, Plappert, Tworek, Hilton, Nakano, Hesse and Schulman}]{Cobbe2021}
\bibinfo{author}{Cobbe, K.}, \bibinfo{author}{Kosaraju, V.}, \bibinfo{author}{Bavarian, M.}, \bibinfo{author}{Chen, M.}, \bibinfo{author}{Jun, H.}, \bibinfo{author}{Kaiser, L.}, \bibinfo{author}{Plappert, M.}, \bibinfo{author}{Tworek, J.}, \bibinfo{author}{Hilton, J.}, \bibinfo{author}{Nakano, R.}, \bibinfo{author}{Hesse, C.}, \bibinfo{author}{Schulman, J.}, \bibinfo{year}{2021}.
\newblock \bibinfo{title}{Training {{Verifiers}} to {{Solve Math Word Problems}}}.
\newblock \URLprefix \url{http://arxiv.org/abs/2110.14168}, \DOIprefix\doi{10.48550/arXiv.2110.14168}, \href{http://arxiv.org/abs/2110.14168}{{\tt arXiv:2110.14168}}.
%Type = Misc
\bibitem[{Crothers et~al.(2023)Crothers, Japkowicz and Viktor}]{Crothers2023}
\bibinfo{author}{Crothers, E.}, \bibinfo{author}{Japkowicz, N.}, \bibinfo{author}{Viktor, H.}, \bibinfo{year}{2023}.
\newblock \bibinfo{title}{Machine {{Generated Text}}: {{A Comprehensive Survey}} of {{Threat Models}} and {{Detection Methods}}}.
\newblock \URLprefix \url{http://arxiv.org/abs/2210.07321}, \DOIprefix\doi{10.48550/arXiv.2210.07321}, \href{http://arxiv.org/abs/2210.07321}{{\tt arXiv:2210.07321}}.
%Type = Misc
\bibitem[{Cui et~al.(2023)Cui, Ma, Cao, Ye and Wang}]{Cui2023}
\bibinfo{author}{Cui, C.}, \bibinfo{author}{Ma, Y.}, \bibinfo{author}{Cao, X.}, \bibinfo{author}{Ye, W.}, \bibinfo{author}{Wang, Z.}, \bibinfo{year}{2023}.
\newblock \bibinfo{title}{Receive, {{Reason}}, and {{React}}: {{Drive}} as {{You Say}} with {{Large Language Models}} in {{Autonomous Vehicles}}}.
\newblock \URLprefix \url{http://arxiv.org/abs/2310.08034}, \DOIprefix\doi{10.48550/arXiv.2310.08034}, \href{http://arxiv.org/abs/2310.08034}{{\tt arXiv:2310.08034}}.
%Type = Book
\bibitem[{Da et~al.(2023)Da, Gao, Mei and Wei}]{Da2023}
\bibinfo{author}{Da, L.}, \bibinfo{author}{Gao, M.}, \bibinfo{author}{Mei, H.}, \bibinfo{author}{Wei, H.}, \bibinfo{year}{2023}.
\newblock \bibinfo{title}{{{LLM Powered Sim-to-real Transfer}} for {{Traffic Signal Control}}}.
\newblock \DOIprefix\doi{10.48550/arXiv.2308.14284}.
%Type = Article
\bibitem[{Devunuri and Lehe(2024a)}]{Devunuri2024b}
\bibinfo{author}{Devunuri, S.}, \bibinfo{author}{Lehe, L.}, \bibinfo{year}{2024}a.
\newblock \bibinfo{title}{{{GTFS Segments}}: {{A Fast}} and {{Efficient Library}} to {{Generate Bus Stop Spacings}}}.
\newblock \bibinfo{journal}{Journal of Open Source Software} \bibinfo{volume}{9}, \bibinfo{pages}{6306}.
\newblock \URLprefix \url{https://joss.theoj.org/papers/10.21105/joss.06306}, \DOIprefix\doi{10.21105/joss.06306}.
%Type = Article
\bibitem[{Devunuri and Lehe(2024b)}]{Devunuri2024c}
\bibinfo{author}{Devunuri, S.}, \bibinfo{author}{Lehe, L.}, \bibinfo{year}{2024}b.
\newblock \bibinfo{title}{A {{Survey}} of {{Errors}} in {{GTFS Static Feeds}} from the {{United States}}}.
\newblock \bibinfo{journal}{Findings} \URLprefix \url{https://findingspress.org/article/116694-a-survey-of-errors-in-gtfs-static-feeds-from-the-united-states}, \DOIprefix\doi{10.32866/001c.116694}.
%Type = Article
\bibitem[{Devunuri et~al.(2024a)Devunuri, Lehe, Qiam, Pandey and Monzer}]{Devunuri2024d}
\bibinfo{author}{Devunuri, S.}, \bibinfo{author}{Lehe, L.J.}, \bibinfo{author}{Qiam, S.}, \bibinfo{author}{Pandey, A.}, \bibinfo{author}{Monzer, D.}, \bibinfo{year}{2024}a.
\newblock \bibinfo{title}{Bus stop spacing statistics: {{Theory}} and evidence}.
\newblock \bibinfo{journal}{Journal of Public Transportation} \bibinfo{volume}{26}, \bibinfo{pages}{100083}.
\newblock \URLprefix \url{https://www.sciencedirect.com/science/article/pii/S1077291X24000031}, \DOIprefix\doi{10.1016/j.jpubtr.2024.100083}.
%Type = Article
\bibitem[{Devunuri et~al.(2024b)Devunuri, Qiam and Lehe}]{Devunuri2024f}
\bibinfo{author}{Devunuri, S.}, \bibinfo{author}{Qiam, S.}, \bibinfo{author}{Lehe, L.J.}, \bibinfo{year}{2024}b.
\newblock \bibinfo{title}{{{ChatGPT}} for {{GTFS}}: Benchmarking {{LLMs}} on {{GTFS}} semantics... and retrieval}.
\newblock \bibinfo{journal}{Public Transport} \URLprefix \url{https://doi.org/10.1007/s12469-024-00354-x}, \DOIprefix\doi{10.1007/s12469-024-00354-x}.
%Type = Article
\bibitem[{Fayyaz et~al.(2017)Fayyaz, Liu and Zhang}]{Fayyaz2017}
\bibinfo{author}{Fayyaz, K.}, \bibinfo{author}{Liu, X.C.}, \bibinfo{author}{Zhang, G.}, \bibinfo{year}{2017}.
\newblock \bibinfo{title}{An efficient {{General Transit Feed Specification}} ({{GTFS}}) enabled algorithm for dynamic transit accessibility analysis}.
\newblock \bibinfo{journal}{PLOS ONE} \bibinfo{volume}{12}, \bibinfo{pages}{e0185333}.
\newblock \URLprefix \url{https://journals.plos.org/plosone/article?id=10.1371/journal.pone.0185333}, \DOIprefix\doi{10.1371/journal.pone.0185333}.
%Type = Misc
\bibitem[{Feng and Chen(2024)}]{Feng2024}
\bibinfo{author}{Feng, S.}, \bibinfo{author}{Chen, C.}, \bibinfo{year}{2024}.
\newblock \bibinfo{title}{Prompting {{Is All You Need}}: {{Automated Android Bug Replay}} with {{Large Language Models}}}.
\newblock \URLprefix \url{http://arxiv.org/abs/2306.01987}, \DOIprefix\doi{10.48550/arXiv.2306.01987}, \href{http://arxiv.org/abs/2306.01987}{{\tt arXiv:2306.01987}}.
%Type = Article
\bibitem[{Fortin et~al.(2016)Fortin, Morency and Tr{\'e}panier}]{Fortin2016}
\bibinfo{author}{Fortin, P.}, \bibinfo{author}{Morency, C.}, \bibinfo{author}{Tr{\'e}panier, M.}, \bibinfo{year}{2016}.
\newblock \bibinfo{title}{Innovative {{GTFS Data Application}} for {{Transit Network Analysis Using}} a {{Graph-Oriented Method}}}.
\newblock \bibinfo{journal}{Journal of Public Transportation} \bibinfo{volume}{19}, \bibinfo{pages}{18--37}.
\newblock \URLprefix \url{https://www.sciencedirect.com/science/article/pii/S1077291X22001151}, \DOIprefix\doi{10.5038/2375-0901.19.4.2}.
%Type = Misc
\bibitem[{Fu et~al.(2023)Fu, Li, Wen, Dou, Cai, Shi and Qiao}]{Fu2023}
\bibinfo{author}{Fu, D.}, \bibinfo{author}{Li, X.}, \bibinfo{author}{Wen, L.}, \bibinfo{author}{Dou, M.}, \bibinfo{author}{Cai, P.}, \bibinfo{author}{Shi, B.}, \bibinfo{author}{Qiao, Y.}, \bibinfo{year}{2023}.
\newblock \bibinfo{title}{Drive {{Like}} a {{Human}}: {{Rethinking Autonomous Driving}} with {{Large Language Models}}}.
\newblock \URLprefix \url{http://arxiv.org/abs/2307.07162}, \DOIprefix\doi{10.48550/arXiv.2307.07162}, \href{http://arxiv.org/abs/2307.07162}{{\tt arXiv:2307.07162}}.
%Type = Misc
\bibitem[{Gao et~al.(2024)Gao, Xiong, Gao, Jia, Pan, Bi, Dai, Sun, Wang and Wang}]{Gao2024}
\bibinfo{author}{Gao, Y.}, \bibinfo{author}{Xiong, Y.}, \bibinfo{author}{Gao, X.}, \bibinfo{author}{Jia, K.}, \bibinfo{author}{Pan, J.}, \bibinfo{author}{Bi, Y.}, \bibinfo{author}{Dai, Y.}, \bibinfo{author}{Sun, J.}, \bibinfo{author}{Wang, M.}, \bibinfo{author}{Wang, H.}, \bibinfo{year}{2024}.
\newblock \bibinfo{title}{Retrieval-{{Augmented Generation}} for {{Large Language Models}}: {{A Survey}}}.
\newblock \URLprefix \url{http://arxiv.org/abs/2312.10997}, \DOIprefix\doi{10.48550/arXiv.2312.10997}, \href{http://arxiv.org/abs/2312.10997}{{\tt arXiv:2312.10997}}.
%Type = Misc
\bibitem[{Haluptzok et~al.(2023)Haluptzok, Bowers and Kalai}]{Haluptzok2023}
\bibinfo{author}{Haluptzok, P.}, \bibinfo{author}{Bowers, M.}, \bibinfo{author}{Kalai, A.T.}, \bibinfo{year}{2023}.
\newblock \bibinfo{title}{Language {{Models Can Teach Themselves}} to {{Program Better}}}.
\newblock \URLprefix \url{http://arxiv.org/abs/2207.14502}, \href{http://arxiv.org/abs/2207.14502}{{\tt arXiv:2207.14502}}.
%Type = Misc
\bibitem[{Hendrycks et~al.(2021)Hendrycks, Burns, Basart, Zou, Mazeika, Song and Steinhardt}]{Hendrycks2021}
\bibinfo{author}{Hendrycks, D.}, \bibinfo{author}{Burns, C.}, \bibinfo{author}{Basart, S.}, \bibinfo{author}{Zou, A.}, \bibinfo{author}{Mazeika, M.}, \bibinfo{author}{Song, D.}, \bibinfo{author}{Steinhardt, J.}, \bibinfo{year}{2021}.
\newblock \bibinfo{title}{Measuring {{Massive Multitask Language Understanding}}}.
\newblock \URLprefix \url{http://arxiv.org/abs/2009.03300}, \DOIprefix\doi{10.48550/arXiv.2009.03300}, \href{http://arxiv.org/abs/2009.03300}{{\tt arXiv:2009.03300}}.
%Type = Misc
\bibitem[{Hsieh et~al.(2024)Hsieh, Sun, Kriman, Acharya, Rekesh, Jia, Zhang and Ginsburg}]{Hsieh2024}
\bibinfo{author}{Hsieh, C.P.}, \bibinfo{author}{Sun, S.}, \bibinfo{author}{Kriman, S.}, \bibinfo{author}{Acharya, S.}, \bibinfo{author}{Rekesh, D.}, \bibinfo{author}{Jia, F.}, \bibinfo{author}{Zhang, Y.}, \bibinfo{author}{Ginsburg, B.}, \bibinfo{year}{2024}.
\newblock \bibinfo{title}{{{RULER}}: {{What}}'s the {{Real Context Size}} of {{Your Long-Context Language Models}}?}
\newblock \URLprefix \url{http://arxiv.org/abs/2404.06654}, \DOIprefix\doi{10.48550/arXiv.2404.06654}, \href{http://arxiv.org/abs/2404.06654}{{\tt arXiv:2404.06654}}.
%Type = Misc
\bibitem[{Khatry et~al.(2023)Khatry, Cahoon, Henkel, Deep, Emani, Floratou, Gulwani, Le, Raza, Shi, Singh and Tiwari}]{Khatry2023}
\bibinfo{author}{Khatry, A.}, \bibinfo{author}{Cahoon, J.}, \bibinfo{author}{Henkel, J.}, \bibinfo{author}{Deep, S.}, \bibinfo{author}{Emani, V.}, \bibinfo{author}{Floratou, A.}, \bibinfo{author}{Gulwani, S.}, \bibinfo{author}{Le, V.}, \bibinfo{author}{Raza, M.}, \bibinfo{author}{Shi, S.}, \bibinfo{author}{Singh, M.}, \bibinfo{author}{Tiwari, A.}, \bibinfo{year}{2023}.
\newblock \bibinfo{title}{From {{Words}} to {{Code}}: {{Harnessing Data}} for {{Program Synthesis}} from {{Natural Language}}}.
\newblock \URLprefix \url{http://arxiv.org/abs/2305.01598}, \DOIprefix\doi{10.48550/arXiv.2305.01598}, \href{http://arxiv.org/abs/2305.01598}{{\tt arXiv:2305.01598}}.
%Type = Inproceedings
\bibitem[{Kunama et~al.(2017)Kunama, Worapan, Phithakkitnukoon and Demissie}]{Kunama2017}
\bibinfo{author}{Kunama, N.}, \bibinfo{author}{Worapan, M.}, \bibinfo{author}{Phithakkitnukoon, S.}, \bibinfo{author}{Demissie, M.}, \bibinfo{year}{2017}.
\newblock \bibinfo{title}{{{GTFS-Viz}}: Tool for preprocessing and visualizing {{GTFS}} data}, in: \bibinfo{booktitle}{Proceedings of the 2017 {{ACM International Joint Conference}} on {{Pervasive}} and {{Ubiquitous Computing}} and {{Proceedings}} of the 2017 {{ACM International Symposium}} on {{Wearable Computers}}}, \bibinfo{publisher}{ACM}, \bibinfo{address}{Maui Hawaii}. pp. \bibinfo{pages}{388--396}.
\newblock \URLprefix \url{https://dl.acm.org/doi/10.1145/3123024.3124415}, \DOIprefix\doi{10.1145/3123024.3124415}.
%Type = Misc
\bibitem[{Lewis et~al.(2021)Lewis, Perez, Piktus, Petroni, Karpukhin, Goyal, K{\"u}ttler, Lewis, Yih, Rockt{\"a}schel, Riedel and Kiela}]{Lewis2021}
\bibinfo{author}{Lewis, P.}, \bibinfo{author}{Perez, E.}, \bibinfo{author}{Piktus, A.}, \bibinfo{author}{Petroni, F.}, \bibinfo{author}{Karpukhin, V.}, \bibinfo{author}{Goyal, N.}, \bibinfo{author}{K{\"u}ttler, H.}, \bibinfo{author}{Lewis, M.}, \bibinfo{author}{Yih, W.t.}, \bibinfo{author}{Rockt{\"a}schel, T.}, \bibinfo{author}{Riedel, S.}, \bibinfo{author}{Kiela, D.}, \bibinfo{year}{2021}.
\newblock \bibinfo{title}{Retrieval-{{Augmented Generation}} for {{Knowledge-Intensive NLP Tasks}}}.
\newblock \URLprefix \url{http://arxiv.org/abs/2005.11401}, \DOIprefix\doi{10.48550/arXiv.2005.11401}, \href{http://arxiv.org/abs/2005.11401}{{\tt arXiv:2005.11401}}.
%Type = Misc
\bibitem[{Liu et~al.(2024)Liu, Guo, Gu, King, Han and Brakewood}]{Liu2024}
\bibinfo{author}{Liu, D.}, \bibinfo{author}{Guo, J.}, \bibinfo{author}{Gu, Y.}, \bibinfo{author}{King, M.}, \bibinfo{author}{Han, L.D.}, \bibinfo{author}{Brakewood, C.}, \bibinfo{year}{2024}.
\newblock \bibinfo{title}{{{GTFS2STN}}: {{Analyzing GTFS Transit Data}} by {{Generating Spatiotemporal Transit Network}}}.
\newblock \URLprefix \url{http://arxiv.org/abs/2405.02760}, \href{http://arxiv.org/abs/2405.02760}{{\tt arXiv:2405.02760}}.
%Type = Misc
\bibitem[{Liu et~al.(2023)Liu, Lin, Hewitt, Paranjape, Bevilacqua, Petroni and Liang}]{Liu2023c}
\bibinfo{author}{Liu, N.F.}, \bibinfo{author}{Lin, K.}, \bibinfo{author}{Hewitt, J.}, \bibinfo{author}{Paranjape, A.}, \bibinfo{author}{Bevilacqua, M.}, \bibinfo{author}{Petroni, F.}, \bibinfo{author}{Liang, P.}, \bibinfo{year}{2023}.
\newblock \bibinfo{title}{Lost in the {{Middle}}: {{How Language Models Use Long Contexts}}}.
\newblock \URLprefix \url{http://arxiv.org/abs/2307.03172}, \DOIprefix\doi{10.48550/arXiv.2307.03172}, \href{http://arxiv.org/abs/2307.03172}{{\tt arXiv:2307.03172}}.
%Type = Incollection
\bibitem[{McHugh(2013)}]{McHugh2013}
\bibinfo{author}{McHugh, B.}, \bibinfo{year}{2013}.
\newblock \bibinfo{title}{Pioneering {{Open Data Standards}}: {{The GTFS Story}}}, in: \bibinfo{booktitle}{Beyond Transparency: Open Data and the Future of Civic Innovation}. \bibinfo{publisher}{Code for America Press San Francisco}, pp. \bibinfo{pages}{125--135}.
\newblock \URLprefix \url{https://beyondtransparency.org/part-2/pioneering-open-data-standards-the-gtfs-story/}.
%Type = Misc
\bibitem[{{MobilityData}(2024a)}]{MobilityData2024a}
\bibinfo{author}{{MobilityData}}, \bibinfo{year}{2024}a.
\newblock \bibinfo{title}{General {{Bikeshare Feed Specification}}}.
\newblock \URLprefix \url{https://gbfs.org/}.
%Type = Misc
\bibitem[{{MobilityData}(2024b)}]{MobilityData2024}
\bibinfo{author}{{MobilityData}}, \bibinfo{year}{2024}b.
\newblock \bibinfo{title}{General {{Transit Feed Specification}}}.
\newblock \URLprefix \url{https://gtfs.org/}.
%Type = Article
\bibitem[{Movahedi and Choi(2024)}]{Movahedi2024}
\bibinfo{author}{Movahedi, M.}, \bibinfo{author}{Choi, J.}, \bibinfo{year}{2024}.
\newblock \bibinfo{title}{The {{Crossroads}} of {{LLM}} and {{Traffic Control}}: {{A Study}} on {{Large Language Models}} in {{Adaptive Traffic Signal Control}}}.
\newblock \bibinfo{journal}{IEEE Transactions on Intelligent Transportation Systems} , \bibinfo{pages}{1--16}\URLprefix \url{https://ieeexplore.ieee.org/document/10768207}, \DOIprefix\doi{10.1109/TITS.2024.3498735}.
%Type = Misc
\bibitem[{Mumtarin et~al.(2023)Mumtarin, Chowdhury and Wood}]{Mumtarin2023}
\bibinfo{author}{Mumtarin, M.}, \bibinfo{author}{Chowdhury, M.S.}, \bibinfo{author}{Wood, J.}, \bibinfo{year}{2023}.
\newblock \bibinfo{title}{Large {{Language Models}} in {{Analyzing Crash Narratives}} -- {{A Comparative Study}} of {{ChatGPT}}, {{BARD}} and {{GPT-4}}}.
\newblock \URLprefix \url{http://arxiv.org/abs/2308.13563}, \DOIprefix\doi{10.48550/arXiv.2308.13563}, \href{http://arxiv.org/abs/2308.13563}{{\tt arXiv:2308.13563}}.
%Type = Book
\bibitem[{Oliveira et~al.(2024)Oliveira, Espadoto, Hirata~Jr, Damaceno and Cesar}]{Oliveira2024}
\bibinfo{author}{Oliveira, A.}, \bibinfo{author}{Espadoto, M.}, \bibinfo{author}{Hirata~Jr, R.}, \bibinfo{author}{Damaceno, R.}, \bibinfo{author}{Cesar, R.}, \bibinfo{year}{2024}.
\newblock \bibinfo{title}{Towards a {{Method}} for {{Evaluating Bus Stop Infrastructure}} with {{Street Level Images}} and {{Large Language Models}}}.
%Type = Misc
\bibitem[{OpenAI et~al.(2024)OpenAI, Achiam, Adler, Agarwal, Ahmad, Akkaya, Aleman, Almeida, Altenschmidt, Altman, Anadkat, Avila, Babuschkin, Balaji, Balcom, Baltescu, Bao, Bavarian, Belgum, Bello, Berdine, {Bernadett-Shapiro}, Berner, Bogdonoff, Boiko, Boyd, Brakman, Brockman, Brooks, Brundage, Button, Cai, Campbell, Cann, Carey, Carlson, Carmichael, Chan, Chang, Chantzis, Chen, Chen, Chen, Chen, Chen, Chess, Cho, Chu, Chung, Cummings, Currier, Dai, Decareaux, Degry, Deutsch, Deville, Dhar, Dohan, Dowling, Dunning, Ecoffet, Eleti, Eloundou, Farhi, Fedus, Felix, Fishman, Forte, Fulford, Gao, Georges, Gibson, Goel, Gogineni, Goh, {Gontijo-Lopes}, Gordon, Grafstein, Gray, Greene, Gross, Gu, Guo, Hallacy, Han, Harris, He, Heaton, Heidecke, Hesse, Hickey, Hickey, Hoeschele, Houghton, Hsu, Hu, Hu, Huizinga, Jain, Jain, Jang, Jiang, Jiang, Jin, Jin, Jomoto, Jonn, Jun, Kaftan, Kaiser, Kamali, Kanitscheider, Keskar, Khan, Kilpatrick, Kim, Kim, Kim, Kirchner, Kiros, Knight, Kokotajlo, Kondraciuk, Kondrich, Konstantinidis, Kosic, Krueger, Kuo, Lampe, Lan, Lee, Leike, Leung, Levy, Li, Lim, Lin, Lin, Litwin, Lopez, Lowe, Lue, Makanju, Malfacini, Manning, Markov, Markovski, Martin, Mayer, Mayne, McGrew, McKinney, McLeavey, McMillan, McNeil, Medina, Mehta, Menick, Metz, Mishchenko, Mishkin, Monaco, Morikawa, Mossing, Mu, Murati, Murk, M{\'e}ly, Nair, Nakano, Nayak, Neelakantan, Ngo, Noh, Ouyang, O'Keefe, Pachocki, Paino, Palermo, Pantuliano, Parascandolo, Parish, Parparita, Passos, Pavlov, Peng, Perelman, Peres, Petrov, Pinto, Michael, Pokorny, Pokrass, Pong, Powell, Power, Power, Proehl, Puri, Radford, Rae, Ramesh, Raymond, Real, Rimbach, Ross, Rotsted, Roussez, Ryder, Saltarelli, Sanders, Santurkar, Sastry, Schmidt, Schnurr, Schulman, Selsam, Sheppard, Sherbakov, Shieh, Shoker, Shyam, Sidor, Sigler, Simens, Sitkin, Slama, Sohl, Sokolowsky, Song, Staudacher, Such, Summers, Sutskever, Tang, Tezak, Thompson, Tillet, Tootoonchian, Tseng, Tuggle, Turley, Tworek, Uribe, Vallone, Vijayvergiya, Voss, Wainwright, Wang, Wang, Wang, Ward, Wei, Weinmann, Welihinda, Welinder, Weng, Weng, Wiethoff, Willner, Winter, Wolrich, Wong, Workman, Wu, Wu, Wu, Xiao, Xu, Yoo, Yu, Yuan, Zaremba, Zellers, Zhang, Zhang, Zhao, Zheng, Zhuang, Zhuk and Zoph}]{OpenAI2024}
\bibinfo{author}{OpenAI}, \bibinfo{author}{Achiam, J.}, \bibinfo{author}{Adler, S.}, \bibinfo{author}{Agarwal, S.}, \bibinfo{author}{Ahmad, L.}, \bibinfo{author}{Akkaya, I.}, \bibinfo{author}{Aleman, F.L.}, \bibinfo{author}{Almeida, D.}, \bibinfo{author}{Altenschmidt, J.}, \bibinfo{author}{Altman, S.}, \bibinfo{author}{Anadkat, S.}, \bibinfo{author}{Avila, R.}, \bibinfo{author}{Babuschkin, I.}, \bibinfo{author}{Balaji, S.}, \bibinfo{author}{Balcom, V.}, \bibinfo{author}{Baltescu, P.}, \bibinfo{author}{Bao, H.}, \bibinfo{author}{Bavarian, M.}, \bibinfo{author}{Belgum, J.}, \bibinfo{author}{Bello, I.}, \bibinfo{author}{Berdine, J.}, \bibinfo{author}{{Bernadett-Shapiro}, G.}, \bibinfo{author}{Berner, C.}, \bibinfo{author}{Bogdonoff, L.}, \bibinfo{author}{Boiko, O.}, \bibinfo{author}{Boyd, M.}, \bibinfo{author}{Brakman, A.L.}, \bibinfo{author}{Brockman, G.}, \bibinfo{author}{Brooks, T.}, \bibinfo{author}{Brundage, M.}, \bibinfo{author}{Button, K.}, \bibinfo{author}{Cai, T.}, \bibinfo{author}{Campbell, R.}, \bibinfo{author}{Cann, A.}, \bibinfo{author}{Carey, B.}, \bibinfo{author}{Carlson, C.}, \bibinfo{author}{Carmichael, R.}, \bibinfo{author}{Chan, B.}, \bibinfo{author}{Chang, C.}, \bibinfo{author}{Chantzis, F.}, \bibinfo{author}{Chen, D.}, \bibinfo{author}{Chen, S.}, \bibinfo{author}{Chen, R.}, \bibinfo{author}{Chen, J.}, \bibinfo{author}{Chen, M.}, \bibinfo{author}{Chess, B.}, \bibinfo{author}{Cho, C.}, \bibinfo{author}{Chu, C.}, \bibinfo{author}{Chung, H.W.}, \bibinfo{author}{Cummings, D.}, \bibinfo{author}{Currier, J.}, \bibinfo{author}{Dai, Y.}, \bibinfo{author}{Decareaux, C.}, \bibinfo{author}{Degry, T.}, \bibinfo{author}{Deutsch, N.}, \bibinfo{author}{Deville, D.}, \bibinfo{author}{Dhar, A.}, \bibinfo{author}{Dohan, D.}, \bibinfo{author}{Dowling, S.}, \bibinfo{author}{Dunning, S.}, \bibinfo{author}{Ecoffet, A.}, \bibinfo{author}{Eleti, A.}, \bibinfo{author}{Eloundou, T.}, \bibinfo{author}{Farhi, D.}, \bibinfo{author}{Fedus, L.}, \bibinfo{author}{Felix, N.}, \bibinfo{author}{Fishman, S.P.}, \bibinfo{author}{Forte, J.}, \bibinfo{author}{Fulford, I.}, \bibinfo{author}{Gao, L.}, \bibinfo{author}{Georges, E.}, \bibinfo{author}{Gibson, C.}, \bibinfo{author}{Goel, V.}, \bibinfo{author}{Gogineni, T.}, \bibinfo{author}{Goh, G.}, \bibinfo{author}{{Gontijo-Lopes}, R.}, \bibinfo{author}{Gordon, J.}, \bibinfo{author}{Grafstein, M.}, \bibinfo{author}{Gray, S.}, \bibinfo{author}{Greene, R.}, \bibinfo{author}{Gross, J.}, \bibinfo{author}{Gu, S.S.}, \bibinfo{author}{Guo, Y.}, \bibinfo{author}{Hallacy, C.}, \bibinfo{author}{Han, J.}, \bibinfo{author}{Harris, J.}, \bibinfo{author}{He, Y.}, \bibinfo{author}{Heaton, M.}, \bibinfo{author}{Heidecke, J.}, \bibinfo{author}{Hesse, C.}, \bibinfo{author}{Hickey, A.}, \bibinfo{author}{Hickey, W.}, \bibinfo{author}{Hoeschele, P.}, \bibinfo{author}{Houghton, B.}, \bibinfo{author}{Hsu, K.}, \bibinfo{author}{Hu, S.}, \bibinfo{author}{Hu, X.}, \bibinfo{author}{Huizinga, J.}, \bibinfo{author}{Jain, S.}, \bibinfo{author}{Jain, S.}, \bibinfo{author}{Jang, J.}, \bibinfo{author}{Jiang, A.}, \bibinfo{author}{Jiang, R.}, \bibinfo{author}{Jin, H.}, \bibinfo{author}{Jin, D.}, \bibinfo{author}{Jomoto, S.}, \bibinfo{author}{Jonn, B.}, \bibinfo{author}{Jun, H.}, \bibinfo{author}{Kaftan, T.}, \bibinfo{author}{Kaiser, {\L}.}, \bibinfo{author}{Kamali, A.}, \bibinfo{author}{Kanitscheider, I.}, \bibinfo{author}{Keskar, N.S.}, \bibinfo{author}{Khan, T.}, \bibinfo{author}{Kilpatrick, L.}, \bibinfo{author}{Kim, J.W.}, \bibinfo{author}{Kim, C.}, \bibinfo{author}{Kim, Y.}, \bibinfo{author}{Kirchner, J.H.}, \bibinfo{author}{Kiros, J.}, \bibinfo{author}{Knight, M.}, \bibinfo{author}{Kokotajlo, D.}, \bibinfo{author}{Kondraciuk, {\L}.}, \bibinfo{author}{Kondrich, A.}, \bibinfo{author}{Konstantinidis, A.}, \bibinfo{author}{Kosic, K.}, \bibinfo{author}{Krueger, G.}, \bibinfo{author}{Kuo, V.}, \bibinfo{author}{Lampe, M.}, \bibinfo{author}{Lan, I.}, \bibinfo{author}{Lee, T.}, \bibinfo{author}{Leike, J.}, \bibinfo{author}{Leung, J.}, \bibinfo{author}{Levy, D.}, \bibinfo{author}{Li, C.M.}, \bibinfo{author}{Lim, R.}, \bibinfo{author}{Lin, M.}, \bibinfo{author}{Lin, S.}, \bibinfo{author}{Litwin, M.}, \bibinfo{author}{Lopez, T.}, \bibinfo{author}{Lowe, R.}, \bibinfo{author}{Lue, P.}, \bibinfo{author}{Makanju, A.}, \bibinfo{author}{Malfacini, K.}, \bibinfo{author}{Manning, S.}, \bibinfo{author}{Markov, T.}, \bibinfo{author}{Markovski, Y.}, \bibinfo{author}{Martin, B.}, \bibinfo{author}{Mayer, K.}, \bibinfo{author}{Mayne, A.}, \bibinfo{author}{McGrew, B.}, \bibinfo{author}{McKinney, S.M.}, \bibinfo{author}{McLeavey, C.}, \bibinfo{author}{McMillan, P.}, \bibinfo{author}{McNeil, J.}, \bibinfo{author}{Medina, D.}, \bibinfo{author}{Mehta, A.}, \bibinfo{author}{Menick, J.}, \bibinfo{author}{Metz, L.}, \bibinfo{author}{Mishchenko, A.}, \bibinfo{author}{Mishkin, P.}, \bibinfo{author}{Monaco, V.}, \bibinfo{author}{Morikawa, E.}, \bibinfo{author}{Mossing, D.}, \bibinfo{author}{Mu, T.}, \bibinfo{author}{Murati, M.}, \bibinfo{author}{Murk, O.}, \bibinfo{author}{M{\'e}ly, D.}, \bibinfo{author}{Nair, A.}, \bibinfo{author}{Nakano, R.}, \bibinfo{author}{Nayak, R.}, \bibinfo{author}{Neelakantan, A.}, \bibinfo{author}{Ngo, R.}, \bibinfo{author}{Noh, H.}, \bibinfo{author}{Ouyang, L.}, \bibinfo{author}{O'Keefe, C.}, \bibinfo{author}{Pachocki, J.}, \bibinfo{author}{Paino, A.}, \bibinfo{author}{Palermo, J.}, \bibinfo{author}{Pantuliano, A.}, \bibinfo{author}{Parascandolo, G.}, \bibinfo{author}{Parish, J.}, \bibinfo{author}{Parparita, E.}, \bibinfo{author}{Passos, A.}, \bibinfo{author}{Pavlov, M.}, \bibinfo{author}{Peng, A.}, \bibinfo{author}{Perelman, A.}, \bibinfo{author}{Peres, F.d.A.B.}, \bibinfo{author}{Petrov, M.}, \bibinfo{author}{Pinto, H.P.d.O.}, \bibinfo{author}{Michael}, \bibinfo{author}{Pokorny}, \bibinfo{author}{Pokrass, M.}, \bibinfo{author}{Pong, V.H.}, \bibinfo{author}{Powell, T.}, \bibinfo{author}{Power, A.}, \bibinfo{author}{Power, B.}, \bibinfo{author}{Proehl, E.}, \bibinfo{author}{Puri, R.}, \bibinfo{author}{Radford, A.}, \bibinfo{author}{Rae, J.}, \bibinfo{author}{Ramesh, A.}, \bibinfo{author}{Raymond, C.}, \bibinfo{author}{Real, F.}, \bibinfo{author}{Rimbach, K.}, \bibinfo{author}{Ross, C.}, \bibinfo{author}{Rotsted, B.}, \bibinfo{author}{Roussez, H.}, \bibinfo{author}{Ryder, N.}, \bibinfo{author}{Saltarelli, M.}, \bibinfo{author}{Sanders, T.}, \bibinfo{author}{Santurkar, S.}, \bibinfo{author}{Sastry, G.}, \bibinfo{author}{Schmidt, H.}, \bibinfo{author}{Schnurr, D.}, \bibinfo{author}{Schulman, J.}, \bibinfo{author}{Selsam, D.}, \bibinfo{author}{Sheppard, K.}, \bibinfo{author}{Sherbakov, T.}, \bibinfo{author}{Shieh, J.}, \bibinfo{author}{Shoker, S.}, \bibinfo{author}{Shyam, P.}, \bibinfo{author}{Sidor, S.}, \bibinfo{author}{Sigler, E.}, \bibinfo{author}{Simens, M.}, \bibinfo{author}{Sitkin, J.}, \bibinfo{author}{Slama, K.}, \bibinfo{author}{Sohl, I.}, \bibinfo{author}{Sokolowsky, B.}, \bibinfo{author}{Song, Y.}, \bibinfo{author}{Staudacher, N.}, \bibinfo{author}{Such, F.P.}, \bibinfo{author}{Summers, N.}, \bibinfo{author}{Sutskever, I.}, \bibinfo{author}{Tang, J.}, \bibinfo{author}{Tezak, N.}, \bibinfo{author}{Thompson, M.B.}, \bibinfo{author}{Tillet, P.}, \bibinfo{author}{Tootoonchian, A.}, \bibinfo{author}{Tseng, E.}, \bibinfo{author}{Tuggle, P.}, \bibinfo{author}{Turley, N.}, \bibinfo{author}{Tworek, J.}, \bibinfo{author}{Uribe, J.F.C.}, \bibinfo{author}{Vallone, A.}, \bibinfo{author}{Vijayvergiya, A.}, \bibinfo{author}{Voss, C.}, \bibinfo{author}{Wainwright, C.}, \bibinfo{author}{Wang, J.J.}, \bibinfo{author}{Wang, A.}, \bibinfo{author}{Wang, B.}, \bibinfo{author}{Ward, J.}, \bibinfo{author}{Wei, J.}, \bibinfo{author}{Weinmann, C.J.}, \bibinfo{author}{Welihinda, A.}, \bibinfo{author}{Welinder, P.}, \bibinfo{author}{Weng, J.}, \bibinfo{author}{Weng, L.}, \bibinfo{author}{Wiethoff, M.}, \bibinfo{author}{Willner, D.}, \bibinfo{author}{Winter, C.}, \bibinfo{author}{Wolrich, S.}, \bibinfo{author}{Wong, H.}, \bibinfo{author}{Workman, L.}, \bibinfo{author}{Wu, S.}, \bibinfo{author}{Wu, J.}, \bibinfo{author}{Wu, M.}, \bibinfo{author}{Xiao, K.}, \bibinfo{author}{Xu, T.}, \bibinfo{author}{Yoo, S.}, \bibinfo{author}{Yu, K.}, \bibinfo{author}{Yuan, Q.}, \bibinfo{author}{Zaremba, W.}, \bibinfo{author}{Zellers, R.}, \bibinfo{author}{Zhang, C.}, \bibinfo{author}{Zhang, M.}, \bibinfo{author}{Zhao, S.}, \bibinfo{author}{Zheng, T.}, \bibinfo{author}{Zhuang, J.}, \bibinfo{author}{Zhuk, W.}, \bibinfo{author}{Zoph, B.}, \bibinfo{year}{2024}.
\newblock \bibinfo{title}{{{GPT-4 Technical Report}}}.
\newblock \URLprefix \url{http://arxiv.org/abs/2303.08774}, \DOIprefix\doi{10.48550/arXiv.2303.08774}, \href{http://arxiv.org/abs/2303.08774}{{\tt arXiv:2303.08774}}.
%Type = Article
\bibitem[{Para et~al.(2024)Para, Wirotsasithon, Jundee, Demissie, Sekimoto, Biljecki and Phithakkitnukoon}]{Para2024}
\bibinfo{author}{Para, S.}, \bibinfo{author}{Wirotsasithon, T.}, \bibinfo{author}{Jundee, T.}, \bibinfo{author}{Demissie, M.G.}, \bibinfo{author}{Sekimoto, Y.}, \bibinfo{author}{Biljecki, F.}, \bibinfo{author}{Phithakkitnukoon, S.}, \bibinfo{year}{2024}.
\newblock \bibinfo{title}{{{G2Viz}}: An online tool for visualizing and analyzing a public transit system from {{GTFS}} data}.
\newblock \bibinfo{journal}{Public Transport} \URLprefix \url{https://link.springer.com/10.1007/s12469-024-00362-x}, \DOIprefix\doi{10.1007/s12469-024-00362-x}.
%Type = Article
\bibitem[{Park et~al.(2020)Park, Mount, Liu, Xiao and Miller}]{Park2020}
\bibinfo{author}{Park, Y.}, \bibinfo{author}{Mount, J.}, \bibinfo{author}{Liu, L.}, \bibinfo{author}{Xiao, N.}, \bibinfo{author}{Miller, H.J.}, \bibinfo{year}{2020}.
\newblock \bibinfo{title}{Assessing public transit performance using real-time data: Spatiotemporal patterns of bus operation delays in {{Columbus}}, {{Ohio}}, {{USA}}}.
\newblock \bibinfo{journal}{International Journal of Geographical Information Science} \bibinfo{volume}{34}, \bibinfo{pages}{367--392}.
\newblock \URLprefix \url{https://doi.org/10.1080/13658816.2019.1608997}, \DOIprefix\doi{10.1080/13658816.2019.1608997}.
%Type = Misc
\bibitem[{Patel et~al.(2024)Patel, Reddy, Bahdanau and Dasigi}]{Patel2024}
\bibinfo{author}{Patel, A.}, \bibinfo{author}{Reddy, S.}, \bibinfo{author}{Bahdanau, D.}, \bibinfo{author}{Dasigi, P.}, \bibinfo{year}{2024}.
\newblock \bibinfo{title}{Evaluating {{In-Context Learning}} of {{Libraries}} for {{Code Generation}}}.
\newblock \URLprefix \url{http://arxiv.org/abs/2311.09635}, \href{http://arxiv.org/abs/2311.09635}{{\tt arXiv:2311.09635}}.
%Type = Article
\bibitem[{Pereira et~al.(2023)Pereira, Andrade and Vieira}]{Pereira2023}
\bibinfo{author}{Pereira, R.H.M.}, \bibinfo{author}{Andrade, P.R.}, \bibinfo{author}{Vieira, J.P.B.}, \bibinfo{year}{2023}.
\newblock \bibinfo{title}{Exploring the time geography of public transport networks with the gtfs2gps package}.
\newblock \bibinfo{journal}{Journal of Geographical Systems} \bibinfo{volume}{25}, \bibinfo{pages}{453--466}.
\newblock \URLprefix \url{https://doi.org/10.1007/s10109-022-00400-x}, \DOIprefix\doi{10.1007/s10109-022-00400-x}.
%Type = Article
\bibitem[{Pereira et~al.(2021)Pereira, Saraiva, Herszenhut, Braga and Conway}]{Pereira2021}
\bibinfo{author}{Pereira, R.H.M.}, \bibinfo{author}{Saraiva, M.}, \bibinfo{author}{Herszenhut, D.}, \bibinfo{author}{Braga, C.K.V.}, \bibinfo{author}{Conway, M.W.}, \bibinfo{year}{2021}.
\newblock \bibinfo{title}{R5r: {{Rapid Realistic Routing}} on {{Multimodal Transport Networks}} with {{R}}{\textsuperscript{5}} in {{R}}}.
\newblock \bibinfo{journal}{Findings} \URLprefix \url{https://findingspress.org/article/21262-r5r-rapid-realistic-routing-on-multimodal-transport-networks-with-r-5-in-r}, \DOIprefix\doi{10.32866/001c.21262}.
%Type = Article
\bibitem[{Prajapati et~al.(2020)Prajapati, Bhattrai and Bajracharya}]{Prajapati2020}
\bibinfo{author}{Prajapati, A.}, \bibinfo{author}{Bhattrai, N.}, \bibinfo{author}{Bajracharya, T.}, \bibinfo{year}{2020}.
\newblock \bibinfo{title}{Spatio-{{Temporal Analysis Of Public Transportation System Using Static Transit Accessibility Methodological Framework}}} .
%Type = Misc
\bibitem[{Roth(2010)}]{Roth2010}
\bibinfo{author}{Roth, M.}, \bibinfo{year}{2010}.
\newblock \bibinfo{title}{How {{Google}} and {{Portland}}'s {{TriMet Set}} the {{Standard}} for {{Open Transit Data}} - {{Streetsblog San Francisco}}}.
\newblock \URLprefix \url{https://sf.streetsblog.org/2010/01/05/how-google-and-portlands-trimet-set-the-standard-for-open-transit-data}.
%Type = Misc
\bibitem[{Salewski et~al.(2023)Salewski, Alaniz, {Rio-Torto}, Schulz and Akata}]{Salewski2023}
\bibinfo{author}{Salewski, L.}, \bibinfo{author}{Alaniz, S.}, \bibinfo{author}{{Rio-Torto}, I.}, \bibinfo{author}{Schulz, E.}, \bibinfo{author}{Akata, Z.}, \bibinfo{year}{2023}.
\newblock \bibinfo{title}{In-{{Context Impersonation Reveals Large Language Models}}' {{Strengths}} and {{Biases}}}.
\newblock \URLprefix \url{http://arxiv.org/abs/2305.14930}, \DOIprefix\doi{10.48550/arXiv.2305.14930}, \href{http://arxiv.org/abs/2305.14930}{{\tt arXiv:2305.14930}}.
%Type = Misc
\bibitem[{Schick et~al.(2023)Schick, {Dwivedi-Yu}, Dess{\`i}, Raileanu, Lomeli, Zettlemoyer, Cancedda and Scialom}]{Schick2023}
\bibinfo{author}{Schick, T.}, \bibinfo{author}{{Dwivedi-Yu}, J.}, \bibinfo{author}{Dess{\`i}, R.}, \bibinfo{author}{Raileanu, R.}, \bibinfo{author}{Lomeli, M.}, \bibinfo{author}{Zettlemoyer, L.}, \bibinfo{author}{Cancedda, N.}, \bibinfo{author}{Scialom, T.}, \bibinfo{year}{2023}.
\newblock \bibinfo{title}{Toolformer: {{Language Models Can Teach Themselves}} to {{Use Tools}}}.
\newblock \URLprefix \url{http://arxiv.org/abs/2302.04761}, \href{http://arxiv.org/abs/2302.04761}{{\tt arXiv:2302.04761}}.
%Type = Misc
\bibitem[{Syed et~al.(2024)Syed, Light, Guo, Zhang, Qin, Ouyang and Hu}]{Syed2024}
\bibinfo{author}{Syed, U.}, \bibinfo{author}{Light, E.}, \bibinfo{author}{Guo, X.}, \bibinfo{author}{Zhang, H.}, \bibinfo{author}{Qin, L.}, \bibinfo{author}{Ouyang, Y.}, \bibinfo{author}{Hu, B.}, \bibinfo{year}{2024}.
\newblock \bibinfo{title}{Benchmarking the {{Capabilities}} of {{Large Language Models}} in {{Transportation System Engineering}}: {{Accuracy}}, {{Consistency}}, and {{Reasoning Behaviors}}}.
\newblock \URLprefix \url{http://arxiv.org/abs/2408.08302}, \href{http://arxiv.org/abs/2408.08302}{{\tt arXiv:2408.08302}}.
%Type = Misc
\bibitem[{Wang and Shalaby(2024)}]{Wang2024}
\bibinfo{author}{Wang, J.}, \bibinfo{author}{Shalaby, A.}, \bibinfo{year}{2024}.
\newblock \bibinfo{title}{Transit {{Pulse}}: {{Utilizing Social Media}} as a {{Source}} for {{Customer Feedback}} and {{Information Extraction}} with {{Large Language Model}}}.
\newblock \URLprefix \url{http://arxiv.org/abs/2410.15016}, \href{http://arxiv.org/abs/2410.15016}{{\tt arXiv:2410.15016}}.
%Type = Misc
\bibitem[{Xu et~al.(2023)Xu, Yang, Lin, Wang, Zhou, Zhang and Mao}]{Xu2023}
\bibinfo{author}{Xu, B.}, \bibinfo{author}{Yang, A.}, \bibinfo{author}{Lin, J.}, \bibinfo{author}{Wang, Q.}, \bibinfo{author}{Zhou, C.}, \bibinfo{author}{Zhang, Y.}, \bibinfo{author}{Mao, Z.}, \bibinfo{year}{2023}.
\newblock \bibinfo{title}{{{ExpertPrompting}}: {{Instructing Large Language Models}} to be {{Distinguished Experts}}}.
\newblock \URLprefix \url{http://arxiv.org/abs/2305.14688}, \DOIprefix\doi{10.48550/arXiv.2305.14688}, \href{http://arxiv.org/abs/2305.14688}{{\tt arXiv:2305.14688}}.
%Type = Article
\bibitem[{Yan et~al.(2022)Yan, Bejleri and Zhai}]{Yan2022}
\bibinfo{author}{Yan, X.}, \bibinfo{author}{Bejleri, I.}, \bibinfo{author}{Zhai, L.}, \bibinfo{year}{2022}.
\newblock \bibinfo{title}{A spatiotemporal analysis of transit accessibility to low-wage jobs in {{Miami-Dade County}}}.
\newblock \bibinfo{journal}{Journal of Transport Geography} \bibinfo{volume}{98}, \bibinfo{pages}{103218}.
\newblock \URLprefix \url{https://www.sciencedirect.com/science/article/pii/S0966692321002714}, \DOIprefix\doi{10.1016/j.jtrangeo.2021.103218}.
%Type = Misc
\bibitem[{Yao et~al.(2023)Yao, Zhao, Yu, Du, Shafran, Narasimhan and Cao}]{Yao2023}
\bibinfo{author}{Yao, S.}, \bibinfo{author}{Zhao, J.}, \bibinfo{author}{Yu, D.}, \bibinfo{author}{Du, N.}, \bibinfo{author}{Shafran, I.}, \bibinfo{author}{Narasimhan, K.}, \bibinfo{author}{Cao, Y.}, \bibinfo{year}{2023}.
\newblock \bibinfo{title}{{{ReAct}}: {{Synergizing Reasoning}} and {{Acting}} in {{Language Models}}}.
\newblock \URLprefix \url{http://arxiv.org/abs/2210.03629}, \DOIprefix\doi{10.48550/arXiv.2210.03629}, \href{http://arxiv.org/abs/2210.03629}{{\tt arXiv:2210.03629}}.
%Type = Misc
\bibitem[{Zan et~al.(2022)Zan, Chen, Yang, Lin, Kim, Guan, Wang, Chen and Lou}]{Zan2022}
\bibinfo{author}{Zan, D.}, \bibinfo{author}{Chen, B.}, \bibinfo{author}{Yang, D.}, \bibinfo{author}{Lin, Z.}, \bibinfo{author}{Kim, M.}, \bibinfo{author}{Guan, B.}, \bibinfo{author}{Wang, Y.}, \bibinfo{author}{Chen, W.}, \bibinfo{author}{Lou, J.G.}, \bibinfo{year}{2022}.
\newblock \bibinfo{title}{{{CERT}}: {{Continual Pre-Training}} on {{Sketches}} for {{Library-Oriented Code Generation}}}.
\newblock \URLprefix \url{http://arxiv.org/abs/2206.06888}, \href{http://arxiv.org/abs/2206.06888}{{\tt arXiv:2206.06888}}.
%Type = Misc
\bibitem[{Zhang et~al.(2024a)Zhang, Ye, Du, Hu, Li, Yang, Liu, Zhao, Li and Mao}]{Zhang2024}
\bibinfo{author}{Zhang, B.}, \bibinfo{author}{Ye, Y.}, \bibinfo{author}{Du, G.}, \bibinfo{author}{Hu, X.}, \bibinfo{author}{Li, Z.}, \bibinfo{author}{Yang, S.}, \bibinfo{author}{Liu, C.H.}, \bibinfo{author}{Zhao, R.}, \bibinfo{author}{Li, Z.}, \bibinfo{author}{Mao, H.}, \bibinfo{year}{2024}a.
\newblock \bibinfo{title}{Benchmarking the {{Text-to-SQL Capability}} of {{Large Language Models}}: {{A Comprehensive Evaluation}}}.
\newblock \URLprefix \url{http://arxiv.org/abs/2403.02951}, \href{http://arxiv.org/abs/2403.02951}{{\tt arXiv:2403.02951}}.
%Type = Inproceedings
\bibitem[{Zhang et~al.(2024b)Zhang, Zheng, Yue and Wang}]{Zhang2024a}
\bibinfo{author}{Zhang, D.}, \bibinfo{author}{Zheng, H.}, \bibinfo{author}{Yue, W.}, \bibinfo{author}{Wang, X.}, \bibinfo{year}{2024}b.
\newblock \bibinfo{title}{Advancing {{ITS Applications}} with~{{LLMs}}: {{A Survey}} on~{{Traffic Management}}, {{Transportation Safety}}, and~{{Autonomous Driving}}}, in: \bibinfo{editor}{Hu, M.}, \bibinfo{editor}{Cornelis, C.}, \bibinfo{editor}{Zhang, Y.}, \bibinfo{editor}{Lingras, P.}, \bibinfo{editor}{{\'S}l{\k e}zak, D.}, \bibinfo{editor}{Yao, J.} (Eds.), \bibinfo{booktitle}{Rough {{Sets}}}, \bibinfo{publisher}{Springer Nature Switzerland}, \bibinfo{address}{Cham}. pp. \bibinfo{pages}{295--309}.
\newblock \DOIprefix\doi{10.1007/978-3-031-65668-2_20}.
%Type = Misc
\bibitem[{Zhang et~al.(2024c)Zhang, Zhang, Ren, Li and Yang}]{Zhang2024b}
\bibinfo{author}{Zhang, L.}, \bibinfo{author}{Zhang, Y.}, \bibinfo{author}{Ren, K.}, \bibinfo{author}{Li, D.}, \bibinfo{author}{Yang, Y.}, \bibinfo{year}{2024}c.
\newblock \bibinfo{title}{{{MLCopilot}}: {{Unleashing}} the {{Power}} of {{Large Language Models}} in {{Solving Machine Learning Tasks}}}.
\newblock \URLprefix \url{http://arxiv.org/abs/2304.14979}, \href{http://arxiv.org/abs/2304.14979}{{\tt arXiv:2304.14979}}.
%Type = Article
\bibitem[{Zhang et~al.(2024d)Zhang, Fu, Liang, Zhang, Yu, Cai and Yao}]{Zhang2024d}
\bibinfo{author}{Zhang, S.}, \bibinfo{author}{Fu, D.}, \bibinfo{author}{Liang, W.}, \bibinfo{author}{Zhang, Z.}, \bibinfo{author}{Yu, B.}, \bibinfo{author}{Cai, P.}, \bibinfo{author}{Yao, B.}, \bibinfo{year}{2024}d.
\newblock \bibinfo{title}{{{TrafficGPT}}: {{Viewing}}, processing and interacting with traffic foundation models}.
\newblock \bibinfo{journal}{Transport Policy} \bibinfo{volume}{150}, \bibinfo{pages}{95--105}.
\newblock \URLprefix \url{https://linkinghub.elsevier.com/retrieve/pii/S0967070X24000726}, \DOIprefix\doi{10.1016/j.tranpol.2024.03.006}.
%Type = Misc
\bibitem[{Zhang et~al.(2024e)Zhang, Jiang, Han, Chen, Yang and Ren}]{Zhang2024e}
\bibinfo{author}{Zhang, Y.}, \bibinfo{author}{Jiang, Q.}, \bibinfo{author}{Han, X.}, \bibinfo{author}{Chen, N.}, \bibinfo{author}{Yang, Y.}, \bibinfo{author}{Ren, K.}, \bibinfo{year}{2024}e.
\newblock \bibinfo{title}{Benchmarking {{Data Science Agents}}}.
\newblock \URLprefix \url{http://arxiv.org/abs/2402.17168}, \href{http://arxiv.org/abs/2402.17168}{{\tt arXiv:2402.17168}}.
%Type = Misc
\bibitem[{Zhang et~al.(2024f)Zhang, Sun, Wang, Nie, Ma, Sun and Li}]{Zhang2024f}
\bibinfo{author}{Zhang, Z.}, \bibinfo{author}{Sun, Y.}, \bibinfo{author}{Wang, Z.}, \bibinfo{author}{Nie, Y.}, \bibinfo{author}{Ma, X.}, \bibinfo{author}{Sun, P.}, \bibinfo{author}{Li, R.}, \bibinfo{year}{2024}f.
\newblock \bibinfo{title}{Large {{Language Models}} for {{Mobility}} in {{Transportation Systems}}: {{A Survey}} on {{Forecasting Tasks}}}.
\newblock \URLprefix \url{http://arxiv.org/abs/2405.02357}, \DOIprefix\doi{10.48550/arXiv.2405.02357}, \href{http://arxiv.org/abs/2405.02357}{{\tt arXiv:2405.02357}}.
%Type = Misc
\bibitem[{Zheng et~al.(2023)Zheng, {Abdel-Aty}, Wang, Wang and Ding}]{Zheng2023}
\bibinfo{author}{Zheng, O.}, \bibinfo{author}{{Abdel-Aty}, M.}, \bibinfo{author}{Wang, D.}, \bibinfo{author}{Wang, Z.}, \bibinfo{author}{Ding, S.}, \bibinfo{year}{2023}.
\newblock \bibinfo{title}{{{ChatGPT}} is on the {{Horizon}}: {{Could}} a {{Large Language Model}} be {{Suitable}} for {{Intelligent Traffic Safety Research}} and {{Applications}}?}
\newblock \URLprefix \url{http://arxiv.org/abs/2303.05382}, \DOIprefix\doi{10.48550/arXiv.2303.05382}, \href{http://arxiv.org/abs/2303.05382}{{\tt arXiv:2303.05382}}.

\end{thebibliography}


% %% For authoryear reference style

% \bibitem[Lamport(1994)]{lamport94}
%   \textit{\LaTeX: a document preparation system},
%   Addison Wesley, Massachusetts,

% \end{thebibliography}
\end{document}